\documentclass{article}

    \usepackage[final, nonatbib]{neurips_2024}

\usepackage[utf8]{inputenc} 
\usepackage[T1]{fontenc}    
\usepackage{hyperref}       
\usepackage{url}            
\usepackage{booktabs}       
\usepackage{amsfonts}       
\usepackage{nicefrac}       
\usepackage{microtype}      
\usepackage{xcolor}         

\usepackage{graphicx}
\usepackage{amsmath} 
\usepackage{cleveref}

\usepackage{multirow}
\usepackage{threeparttable}
\usepackage{colortbl}  
\usepackage{array}   
\usepackage{xspace}
\newcommand{\ie}{{\emph{i.e.}},\xspace}

\newcommand{\etc}{etc.}
\newcommand{\etal}{{\emph{et al.}}}

\definecolor{yzybest}{rgb}{0.96, 0.57, 0.58}
\definecolor{yzysecond}{rgb}{0.98, 0.78, 0.57}
\definecolor{yzythird}{rgb}{1.0, 1.0, 0.56}

\definecolor{gscolor}{rgb}{1.0,0.6,0.0} %

\makeatletter
\newcommand{\printfnsymbol}[1]{%
  \textsuperscript{\@fnsymbol{#1}}%
}
\makeatother

\title{MotionGS: Exploring Explicit Motion Guidance for Deformable 3D Gaussian Splatting}

\author{
Ruijie Zhu$^{}$\thanks{Equal contribution} ~\quad
Yanzhe Liang$^{*}$ ~\quad 
Hanzhi Chang ~\quad 
Jiacheng Deng ~\quad 
\textbf{Jiahao Lu} ~\quad \\ 
\textbf{Wenfei Yang} ~\quad 
\textbf{Tianzhu Zhang} ~\quad 
\textbf{Yongdong Zhang} \\\\
\textbf{University of Science and Technology of China} \\\\
\texttt{\{ruijiezhu, yzliang, changhz, dengjc, lujiahao\}@mail.ustc.edu.cn,}\\
\texttt{\{yangwf, tzzhang, zhyd73\}@ustc.edu.cn}
}

\begin{document}

\maketitle

\begin{abstract}
Dynamic scene reconstruction is a long-term challenge in the field of 3D vision. Recently, the emergence of 3D Gaussian Splatting has provided new insights into this problem. Although subsequent efforts rapidly extend static 3D Gaussian to dynamic scenes, they often lack explicit constraints on object motion, leading to optimization difficulties and performance degradation. 
To address the above issues, we propose a novel deformable 3D Gaussian splatting framework called MotionGS, which explores explicit motion priors to guide the deformation of 3D Gaussians. 
Specifically, we first introduce an optical flow decoupling module that decouples optical flow into camera flow and motion flow, corresponding to camera movement and object motion respectively. Then the motion flow can effectively constrain the deformation of 3D Gaussians, thus simulating the motion of dynamic objects. Additionally, a camera pose refinement module is proposed to alternately optimize 3D Gaussians and camera poses, mitigating the impact of inaccurate camera poses. 
Extensive experiments in the monocular dynamic scenes validate that MotionGS surpasses state-of-the-art methods and exhibits significant superiority in both qualitative and quantitative results.
Project page: \url{https://ruijiezhu94.github.io/MotionGS_page/}.
\end{abstract}

\section{Introduction}
\label{sec:intro}

Dynamic scene reconstruction aims to model the 3D structure and appearance of time-evolving scenes, enabling novel-view synthesis at arbitrary timestamps. It is a crucial task in the field of 3D computer vision, attracting widespread attention from the research community and finding important applications in areas such as virtual/augmented reality and 3D content production. In comparison to static scene reconstruction, dynamic scene reconstruction remains a longstanding open challenge due to the difficulties arising from motion complexity and topology changes.

In recent years, a plethora of dynamic scene reconstruction methods~\cite{pumarola2021d, park2021nerfies,li2021neural,liu2023robust,gao2022monocular, li2023dynibar,cao2023hexplane,shao2023tensor4d} have been proposed based on Neural Radiance Fields (NeRF)~\cite{mildenhall2021nerf}, driving rapid advancements in this field. While these methods exhibit impressive visual quality, their substantial computational overhead impedes their applications in real-time scenarios. Recently, a novel approach called 3D Gaussian Splatting (3DGS)~\cite{kerbl20233d}, has garnered widespread attention in the research community. By introducing explicit 3D Gaussian representation and efficient CUDA-based rasterizer, 3DGS has achieved unprecedented high-quality novel-view synthesis with real-time rendering. 
Subsequent methods~\cite{luiten2023dynamic, li2023spacetime, wu20234d, yang2023gs4d, huang2023sc, lin2023gaussian, yang2023deformable} rapidly extend 3DGS to dynamic scenes, also named 4D scenes. Initially, D-3DGS~\cite{luiten2023dynamic} proposes to iteratively reconstruct the scene frame by frame, but it incurs significant memory overhead. The more straightforward approaches~\cite{wu20234d, yang2023deformable} utilize a deformation field to simulate the motion of objects by moving the 3D Gaussians to their corresponding positions at different time steps. 
Besides, some methods~\cite{li2023spacetime, duan20244d} do not independently model motion but treat space-time as a whole to optimize. 
While these methods effectively extend 3DGS to dynamic scenes, they rely solely on appearance to supervise dynamic scene reconstruction, lacking explicit motion guidance on Gaussian deformation. When object motion is irregular (e.g., sudden movements), the model may encounter optimization difficulties and fall into local optima.

\begin{figure}
   \centering
   \includegraphics[width=\linewidth]{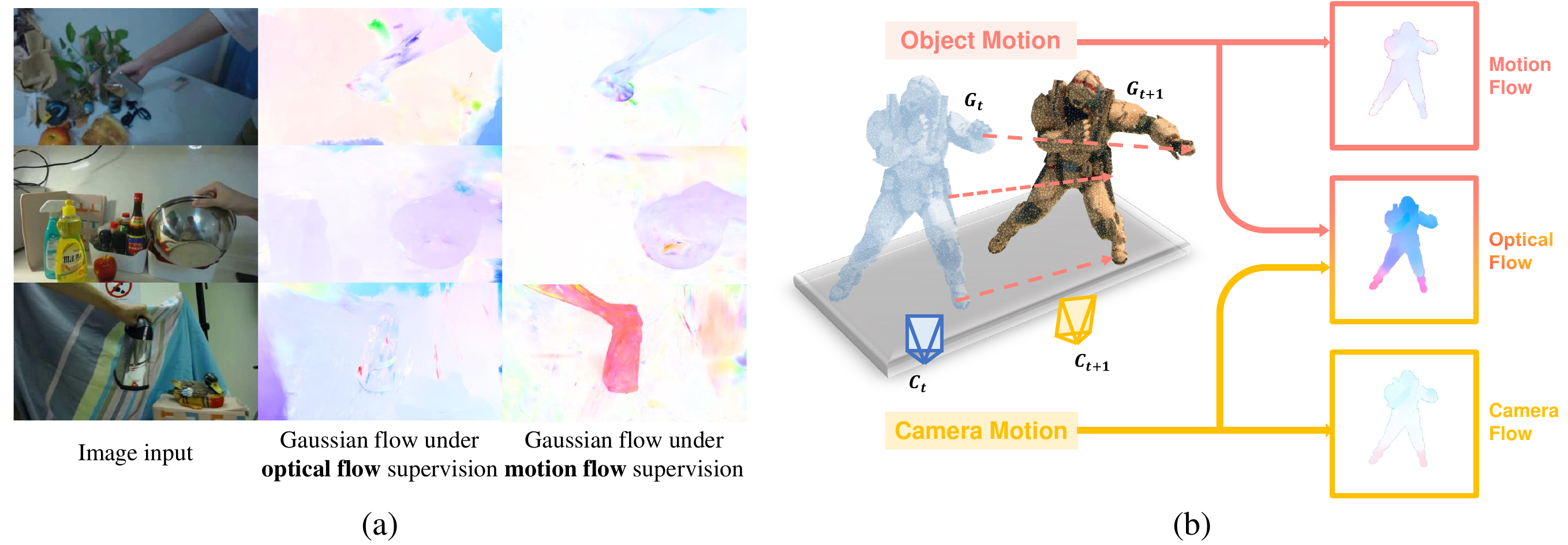}
   \vspace{-1.5em}   
   \caption{
   \textbf{(a) Gaussian flow under different supervision.} 
   We model Gaussian flow under the supervision of optical flow and motion flow respectively. The latter can produce a more direct description of object motion, thereby effectively guiding the deformation of 3D Gaussians.
   \label{fig: moti1}
   \textbf{(b) The decoupling of optical flow.}
   We decouple the optical flow into motion flow which is only related to object motion and camera flow which is only related to camera motion.
   \label{fig: moti2}
   }   
\vspace{-1.3em} 
\end{figure}

Based on the above discussions, we argue that explicit motion guidance is indispensable for the deformation of 3D Gaussians. Benefiting from the advancements in optical flow estimation~\cite{huang2022flowformer, xu2022gmflow}, a natural solution is to utilize an off-the-shelf optical flow network to provide 2D motion priors~\cite{gao2024gaussianflow, guo2024motion}. 
However, the formation of optical flow is affected by both camera motion and object motion, which is not conducive to explicit modeling of object motion. 
Therefore, it is necessary to separate the optical flow related only to the moving object (\ie motion flow) to guide Gaussian deformation more efficiently.
As shown in~\Cref{fig: moti1}(a), directly using optical flow (column~2) to supervise the Gaussian deformation will inevitably include the contribution of static objects to the optical flow, while using motion flow as supervision (column~3) can easily avoid this. 
Besides, the estimated camera pose in dynamic scenes is not always accurate. Due to the lack of geometric consistency between adjacent frames for moving objects, using point correspondences on dynamic objects to calculate the camera pose can lead to erroneous offsets, thereby affecting the optimization of 3DGS.

To address the above issues, we propose a novel deformable 3D Gaussian Splatting framework called MotionGS, which explicitly constrains the deformation of 3D Gaussians by extracting the motion priors from optical flow. Our method includes an optical flow decoupling module and a camera pose refinement module. In the optical flow decoupling module, we decouple the 2D optical flow into camera flow and motion flow, as shown in~\Cref{fig: moti2}(b). The camera flow comes from the camera ego-motion, while the motion flow comes from the motion of dynamic objects. We use the motion flow to directly constrain the deformation of 3D Gaussians (\ie Gaussian flow). Since the calculation of Gaussian flow is directly implemented in the CUDA-based rasterizer, this process is differentiable and efficient. In the camera pose refinement module, we first fix the 3D Gaussians and then utilize photometric consistency loss to backpropagate gradients to camera poses, thereby alternately optimizing 3D Gaussians and camera poses to further enhance the rendering quality.

To sum up, our main contributions are as follows:
\begin{itemize}
    \item We propose a novel deformable 3D Gaussian framework called MotionGS, which provides explicit motion guidance for deformable 3DGS and achieves high-quality dynamic scene reconstruction with real-time rendering.
    \item The proposed optical flow decoupling module effectively separates the flow caused solely by object motion, thereby efficiently supervising the deformation of 3D Gaussians. The proposed pose refinement module alternately optimizes 3DGS and camera poses, reducing reliance on accurate camera poses and further boosting rendering quality.
    \item Extensive experiments have demonstrated the effectiveness of the proposed method. Results on the NeRF-DS and HyperNeRF datasets validate the state-of-the-art performance of our approach in dynamic scene reconstruction.
\end{itemize}

\section{Related Work}
\label{sec:relatedwork}

\subsection{Novel-View Synthesis (NVS)}
Novel view synthesis has been a hot research topic in the field of computer vision and graphics in recent years. NeRF~\cite{mildenhall2021nerf}, which represents 3D scene by neural radiance fields, first achieves high-resolution photorealistic results in this field. Despite many subsequent works~\cite{barron2021mip, barron2022mip, sun2022direct, xu2022point, wang2021ibrnet, muller2022instant, yu2021plenoctrees, zhang2020nerf++, zhu2023tiface} have been proposed to improve its efficiency and quality, NeRF-based methods still struggle to render high-quality images with real-time rendering speed. Recently, by modeling 3D scenes using a set of anisotropic 3D Gaussian with an efficient rasterizer, 3D Gaussian Splatting (3DGS)~\cite{kerbl20233d} has shown remarkable performance with real-time rendering. Compared to NeRF, 3DGS is an explicit 3D scene representation method with better scalability and editability. Therefore, it has been rapidly extended to other 3D vision tasks, including sparse-view reconstruction \cite{li2024dngaussian, chen2024mvsplat, charatan2024pixelsplat, szymanowicz2024splatter_image}, 3D generation~\cite{tang2023dreamgaussian, blattmann2023align, tang2024lgm, wang2023prolificdreamer, xu2024grm}, scene editing~\cite{chen2023gaussianeditor, GaussianEditor, zhou2023feature} and SLAM~\cite{yan2023gs, keetha2024splatam, Matsuki:Murai:etal:CVPR2024, hhuang2024photoslam}.

\subsection{Dynamic Scene Reconstruction}
In recent years, various dynamic scene reconstruction approaches have been proposed, which can be broadly categorized into NeRF-based and 3DGS-based methods. 
NeRF-based works~\cite{du2021neural, pumarola2021d, park2021hypernerf, park2021nerfies, liu2023robust, wang2023tracking, yang2022banmo, gao2022monocular} usually map dynamic scenes to a canonical space and render images based on this 3D canonical space. This kind of 4D scene representation is intuitive but requires a well-reconstructed canonical space. 
Other works propose to use time-varying NeRFs~\cite{li2023dynibar, li2021neural, xian2021space, wang2021neural, gao2021dynamic} or explicit representations~\cite{zhou2024dynpoint,fridovich2023k, cao2023hexplane, shao2023tensor4d, xu20234k4d, wang2023mixed, fang2022fast} to represent and render dynamic scenes. 
However, all these NeRF-based methods require frequent point sampling or MLP queries, suffering from long training and rendering time. 
With the proposal of 3DGS, many works~\cite{lin2023gaussian, gao2024gaussianflow, huang2023sc, luiten2023dynamic, wu20234d, yang2023deformable, li2023spacetime, katsumata2023efficient, yang2023gs4d, lu2024gagaussian, das2023neural, sun20243dgstream} use 3DGS as the fundamental model for 4D scene representation. For instance, D-3DGS~\cite{luiten2023dynamic} models dynamic scenes by allowing the positions and rotation matrixes of 3DGS to change over time. Deformable 3DGS~\cite{yang2023deformable} uses an MLP to model a deformation field based on time and the canonical Gaussian space. 
SC-GS~\cite{huang2023sc} bounds dense 3DGS with sparse control points, calculating the movement of Gaussians in a coarse-to-fine manner. 
Despite they have performed impressive rendering quality in some dynamic scenes, they lack explicit motion guidance to constrain the movement of Gaussian, resulting in degraded performance in more complex dynamic scenes.
Recent works~\cite{gao2024gaussianflow, guo2024motion} compose the movement of 3D points through their corresponding Gaussians, using 2D flow priors to supervise the deformation of 3DGS. 
Inspired by them, we decompose the optical flow to obtain more direct motion supervision, thus achieving higher rendering quality.

\subsection{NVS with Pose Optimization}
Several NVS works~\cite{yen2021inerf, wang2021nerf, xia2022sinerf, lin2021barf, chng2022garf, bian2023nope, fu2023colmap} have noticed that it is difficult to derive precise camera poses of input images in the real world, so they address novel view synthesis together with camera pose optimization. i-NeRF~\cite{yen2021inerf} initially estimates camera poses by matching the input images. Other methods such as NeRFmm~\cite{wang2021nerf} and Nope-NeRF~\cite{bian2023nope} use monocular depth priors as guidance to do the joint optimization of NeRF and camera poses. Recently, CF-3DGS~\cite{fu2023colmap} proposes progressive reconstruction and leverages photometric loss to learn the affine transformation of Gaussians to optimize the camera pose. 
However, these methods are mostly effective only for static scenes and lack support for dynamic scenes.
Motivated by these methods, we aim to extend 3DGS to dynamic scenes with pose optimization, thus boosting the rendering quality and robustness.

\section{Preliminary}
\label{sec:preliminary}

In this section, we briefly introduce the modeling and rendering of 3DGS in~\Cref{sec:preliminary_3DGS} and the deformable extension of 3DGS towards dynamic scene reconstruction in~\Cref{sec:preliminary_Deformable}.

\subsection{3D Gaussian Splatting}
\label{sec:preliminary_3DGS}
As an explicit 3D representation similar to point clouds, 3DGS models the scene with a set of 3D Gaussians.
However, different from point clouds, each 3D Gaussian in the scene has its own opacity $o \in [0,1]$, center position $\mu \in \mathbb{R}^{3\times1}$, and covariance matrix $\Sigma \in \mathbb{R}^{3\times3}$. These properties determine the contribution and influence range of 3D Gaussians on rendering.
For a position $x \in \mathbb{R}^{3\times1}$ in 3D space, the corresponding contribution of a 3D Gaussian on it can be formulated as:
\begin{equation}
\label{formula:gaussian's formula}
    G(x) = o \cdot e^{ -\frac{1}{2} (x-\mu)^{\top} \Sigma^{-1} (x-\mu) }.
\end{equation}
For differentiable optimization, the covariance matrix $\Sigma$ can be decomposed into a scaling matrix $\mathbf{S}$ and a rotation matrix $\mathbf{R}$:
    $\Sigma = \mathbf{R}\mathbf{S}\mathbf{S}^T\mathbf{R}^T$,
where scaling matrix $\mathbf{S} = \text{diag}([s_x, s_y, s_z])$ and rotation matrix $\mathbf{R}$ can be transformed from a quaternion $[r_w, r_x, r_y, r_z]$.
Then the 3D Gaussians can be splatted to a 2D camera plane through differential gaussian splatting.
Specially, given a viewing transform matrix $W$ and the Jacobian matrix $J$ of the affine approximation of the projective transformation, we can obtain the 2D covariance matrix $\Sigma_{\text{2D}}$ through:
    $\Sigma_{\text{2D}} = JW\Sigma W^TJ^T$.
Similarly, we can obtain the 2D center position $\mu_{\text{2D}}$ of 3D Gaussians in camera plane.
Therefore, given a 2D pixel $p$, the rendering contribution of a 3D Gaussian on the viewpoint $W$ can be obtained through a 2D version of~\eqref{formula:gaussian's formula}.
To model the appearance of 3D Gaussians, spherical harmonics (SH) are introduced to define the color $c$.
Finally, for each pixel, the rendering results of 3DGS can be derived by calculating the color contribution of all the related Gaussians. This process is known as $\alpha$-blending:
\begin{equation}
\label{formula: splatting&volume rendering}
    C = \sum_{i}^N c_i \alpha_i \prod_{j=1}^{i-1} (1-\alpha_j),
\end{equation}
where $c_i$, $\alpha_i$ represent the color and density computed from the $i$-th 3D Gaussian.

\subsection{Deformable 3D Gaussian Splatting}
\label{sec:preliminary_Deformable}

To extend 3DGS to dynamic scenes, an intuitive approach is to utilize a learnable deformation field to fit the movement of objects in the real world through Gaussian deformation. This idea originates from NeRF-based methods such as D-NeRF~\cite{pumarola2021d} and has been effectively applied to 3DGS in recent works~\cite{yang2023deformable, wu20234d}.
In these deformable 3DGS methods, a deformation network $\mathcal{D}$ is typically used to model the movement of the center position of 3D Gaussians. 
Additionally, due to the inherent properties of 3D Gaussians, the deformation network $\mathcal{D}$ may also consider the rotation and scaling factors of 3D Gaussians as they vary over time.
Therefore, the deformation of 3D Gaussians can be formulated as:
\begin{equation}
    (\mu + \Delta \mu, r + \Delta r, s + \Delta s) = \mathcal{D}(\mu, r, s, t) ,    
\end{equation}
where $t$ is the timestamp, $\mu, r, s$ are the center position, rotation quaternion and scaling factors of 3D Gaussians, and $\Delta \mu, \Delta r, \Delta s$ are their residuals, respectively.
Due to the various implementations of deformable 3DGS, in this paper we focus solely on the deformation aspect without discussing the other designs and specific differences in these works. 
We select method~\cite{yang2023deformable} as our baseline, leveraging explicit motion guidance and camera pose refinement to further enhance the rendering quality and the robustness in dynamic scenes.

\section{Methodology}
\label{sec:method}

In this section, we first introduce the overall architecture of our approach in~\Cref{sec:method_architecture}. 
Then the optical flow decoupling module is introduced to derive motion guidance for Gaussian deformation in~\Cref{sec:method_flow}.
The camera pose refinement module is introduced to alternately optimize 3D Gaussians and camera poses in~\Cref{sec:method_pose}. Finally, the overall loss function is introduced in~\Cref{sec:method_training}.

\subsection{Overall Architecture}
\label{sec:method_architecture}

\begin{figure}
   \centering
   \includegraphics[width=\linewidth]{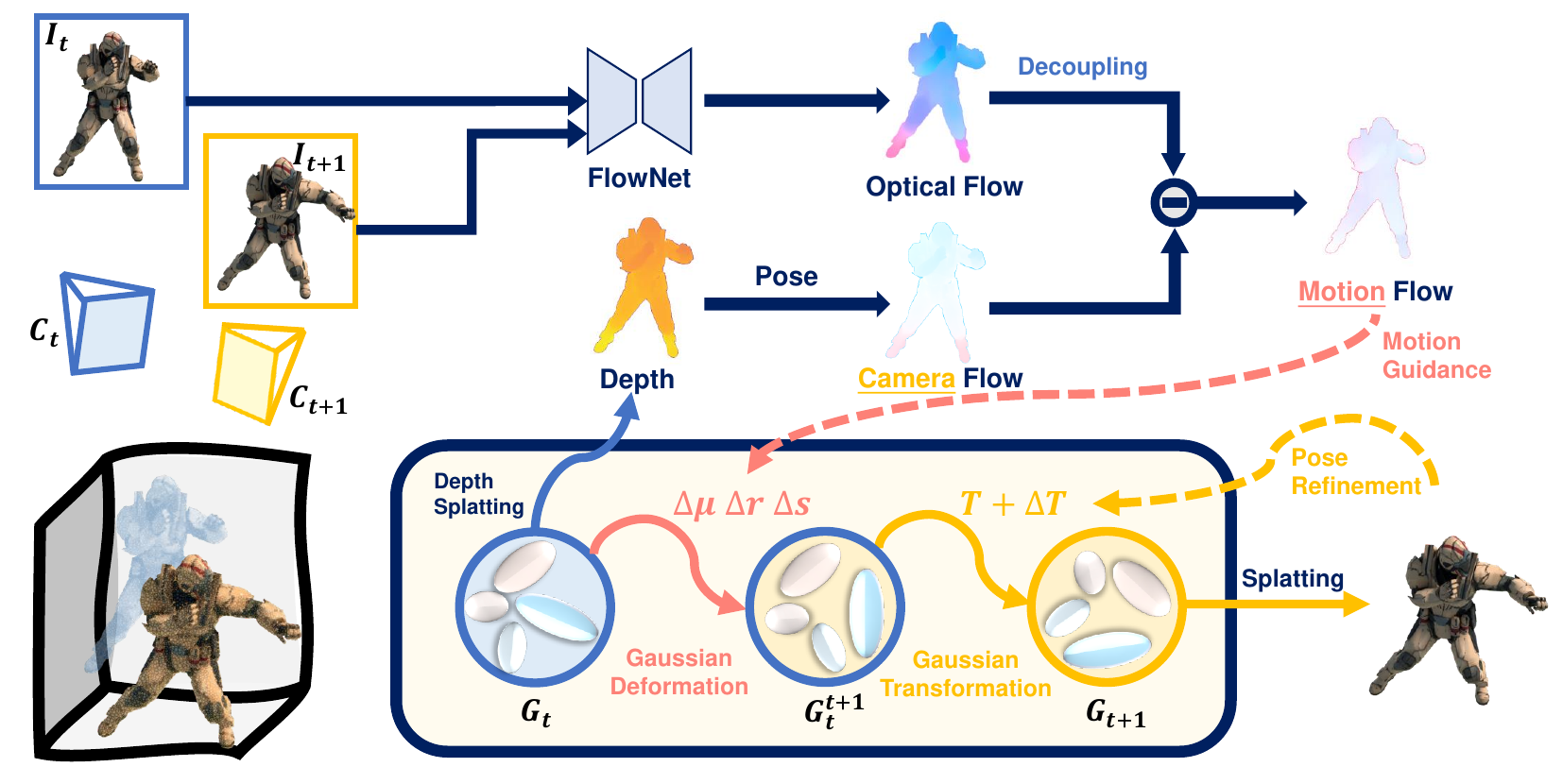}
   \caption{
   \textbf{The overall architecture of MotionGS.} It can be viewed as two data streams: (1) The 2D data stream utilizes the optical flow decoupling module to obtain the motion flow as the 2D motion prior; (2) The 3D data stream involves the deformation and transformation of Gaussians to render the image for the next frame. During training, we alternately optimize 3DGS and camera poses through the camera pose refinement module. }
   \label{fig: method}
   \vspace{-1em}
\end{figure}

The overall architecture of our method is illustrated in~\Cref{fig: method}.
Our method primarily focuses on the reconstruction of monocular dynamic scenes. 
Firstly, following 3DGS~\cite{kerbl20233d}, we initialize camera poses and 3D Gaussians using COLMAP~\cite{schoenberger2016sfm}. 
Given two adjacent frames $I_t$ and $I_{t+1}$, we compute forward optical flow $F_{t \to t+1}$ using an off-the-shelf flow estimation network. 
Meanwhile, we can obtain the rendered depth map $D_t$ of frame $I_t$ at time $t$ through the rasterizer. 
By feeding the depth map $I_t$, camera poses $C_t, C_{t+1}$, and optical flow priors $F_{t \to t+1}$ into the optical flow decoupling module, we can calculate the motion flow $F_{t \to t+1}^M$ solely related to object movement. 
After predicting the deformation of Gaussians through the deformation network $\mathcal{D}$, we obtain the state of 3D Gaussians at time $t+1$ and render the Gaussian flow $F_{t \to t+1}^G$ from time $t$ to $t+1$ under the assumption of a stationary camera viewpoint for the frame $I_t$. 
The motion flow should be consistent with the Gaussian flow, thus providing explicit motion guidance to Gaussian deformation. 
Additionally, since the initialized camera poses may be inaccurate, we add a small residual $\Delta T$ to the relative camera pose $T$. 
Leveraging the proposed camera pose refinement module, we cleverly backpropagate gradients to the camera poses, achieving refinement of the camera poses. 
During training, we alternately optimize 3D Gaussian and camera poses to enhance the rendering quality and robustness in dynamic scenes.

\subsection{Optical Flow Decoupling Module}
\label{sec:method_flow}

To provide explicit motion guidance for the deformation of Gaussians, we first utilize an off-the-shelf optical flow network to predict 2D motion priors. 
Since optical flow is influenced by both camera movement and object motion, we decompose it into camera flow and motion flow as illustrated in the~\Cref{fig: moti2}(b). 
Camera flow represents the optical flow caused solely by camera movement, assuming the objects in the scene remain stationary. 
In contrast, motion flow considers the camera as stationary, capturing only the movement of the objects. 
Essentially, optical flow can be viewed as the vector sum of these two components. 
By decoupling them, we can effectively isolate object motion, providing precise guidance for Gaussian deformation.

\begin{figure}[t]
\begin{minipage}[t]{0.35\linewidth}
   \centering
   \includegraphics[width=\linewidth]{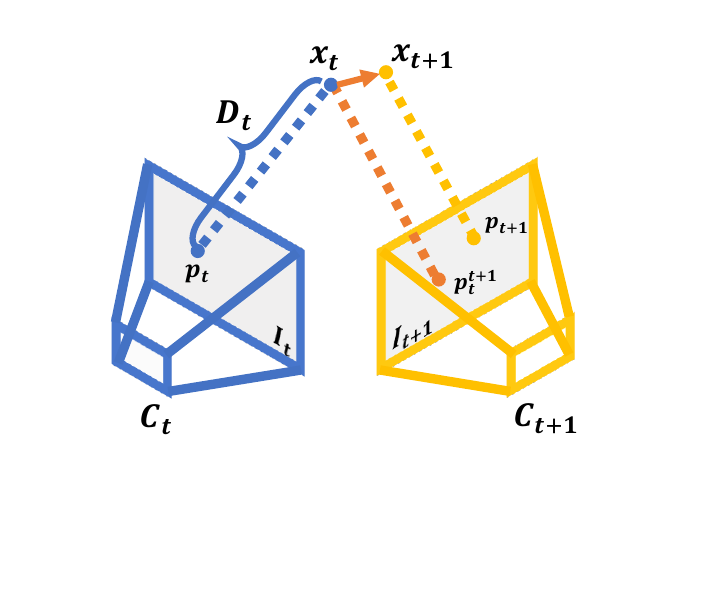}
   \caption{\textbf{Flow calculation.} }
   \label{fig: projection}
   \vspace{-1em}
\end{minipage}
\begin{minipage}[t]{0.65\linewidth}
   \centering
   \includegraphics[width=\linewidth]{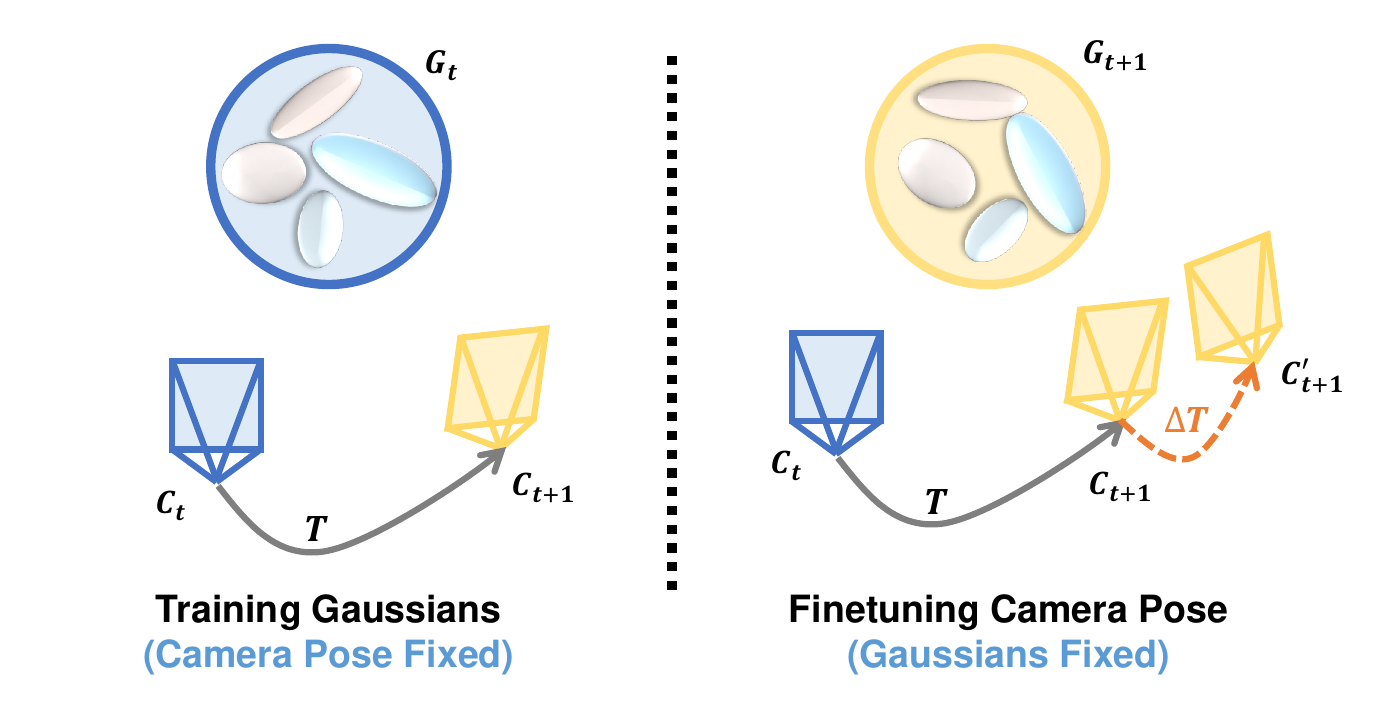}
   \caption{\textbf{Pose refinement on iterative training.} }
   \label{fig: poserefine}
   \vspace{-1em}
\end{minipage}
\end{figure}

\paragraph{Camera flow and motion flow.}
We use a schematic diagram~\Cref{fig: projection} to illustrate the detailed calculation process.
Camera flow can be directly computed from the camera poses and the depth of the current frame. 
Specifically, at the timestamp $t$, we obtain the depth map $D_t$ corresponding to frame $I_t$ directly from 3D Gaussians through the rasterizer. 
Given the intrinsics $K_t$ and extrinsics $T_t$ of camera $C_t$, we can reproject point $p_t$ from frame $I_t$ to 3D space using its depth $D_t$:
\begin{equation}
    x_t = T_t^{-1} K_{t}^{-1} D_t \tilde{p}_t,
\end{equation}
where $\tilde{p}_t$ is the homogeneous coordinate of ${p}_t$.
Assuming $x_t$ does not move, we can obtain the projection $p_t^{t+1}$ of $x_t$ on frame $I_{t+1}$:
\begin{equation}
    p_t^{t+1} = \text{proj}(K_{t+1} T_{t+1} x_t),
\end{equation}
where $K_{t+1}$ and $T_{t+1}$ are the intrinsics and extrinsics of camera $C_{t+1}$, $\text{proj}()$ projects the 3D coordinates to 2D image planes by dividing the last dimension (depth). Then the camera flow can be defined as:
\begin{equation}
    F_{t \to t+1}^C = p_t^{t+1} - p_t,
\end{equation}
which indicates the flow caused solely by camera movement. 
As the point \( x_t \) moves over time, we denote its updated position as \( x_{t+1} \). 
This new point \( x_{t+1} \) is then projected onto frame \( I_{t+1} \) as \( p_{t+1} \). 
Thus, the optical flow $ F_{t \to t+1}$ between two adjacent frame is defined as \( p_{t+1} - p_t \). 
Finally, the motion flow $F_{t \to t+1}^M$ is derived by subtracting the camera flow from the optical flow:
\begin{equation}
    F_{t \to t+1}^M = F_{t \to t+1} - F_{t \to t+1}^C = p_{t+1} - p_{t}^{t+1},
\end{equation}
which also corresponds to the optical flow caused by object movement at a fixed viewpoint.

\paragraph{Gaussion flow.}
To establish a correspondence between Gaussian deformation and motion flow, we need to splat the Gaussian deformation onto the 2D image plane, which is not implemented in the original 3DGS framework.
Inspired by recent work~\cite{gao2024gaussianflow}, we introduce the concept of Gaussian flow, denoted as $F_{t \to t+1}^G$, to describe the 2D projection of Gaussian deformation, and implement it in the CUDA-based rasterizer.
The core idea is to model the contribution of Gaussians to the optical flow by first transforming 3D Gaussians to canonical Gaussian space and then transforming them back to the state at the next time step. 
Please refer to~\Cref{appendix: gaussian flow} for the specific derivation and modeling process of Gaussian flow.
Gao~\etal~\cite{gao2024gaussianflow} computes the deformation of 3D Gaussians from time \( t \) to \( t+1 \) under the transformation of the camera viewpoint from \( C_t \) to \( C_{t+1} \), corresponding to optical flow.
Different from it, our Gaussian flow is designed to match the motion flow, representing the deformation of 3D Gaussians from time \( t \) to \( t+1 \) fixed under the camera viewpoint \( C_{t+1} \). 

\paragraph{Flow loss.}
To effective constrain the Gaussian deformation, we use a $\mathcal{L}_1$ loss between motion flow and Gaussian flow for simplicity:
\begin{equation}
    \mathcal{L}_\text{flow} = \left \| sg(F_{t \to t+1}^M)  - F_{t \to t+1}^G \right \| ,
\end{equation}
where $sg()$ means stop gradient. Note that we also stop the gradients of all variables at time $t$ in the calculation of Gaussian flow for more efficient training.

\paragraph{Discussion.}
The benefits of decoupling the optical flow are evident. Since motion flow is only related to object motion, it can directly provide motion guidance. 
More importantly, in some previous works~\cite{liu2023robust, gao2021dynamic,li2023dynibar}, an off-the-shelf segmentation network is often used to segment out the dynamic objects in the scene (such as humans, animals, cars, \etc). 
However, such masks are only used in their photometric loss to mask out dynamic regions.
In contrast, our motion flow benefits from these dynamic masks more directly. 
By masking static objects with these masks, we can obtain a clear motion flow for supervising Gaussian deformation.
If optical flow is used as motion guidance, this advantage will no longer exist because static objects can also contribute to the optical flow.

\subsection{Camera Pose Refinement Module}
\label{sec:method_pose}

In monocular dynamic scenes, due to the complexity of motion and sparsity of observations, even widely used methods like COLMAP~\cite{schoenberger2016sfm} cannot accurately estimate camera poses. 
Since the optimization of 3DGS requires precise camera poses as input, it often performs poorly in complex dynamic scenes. 
Existing 3DGS-based dynamic scene reconstruction methods rarely take this into account. 
Inspired by pose-free optimization methods for static scene reconstruction~\cite{MatsukiCVPR2024,fu2023colmap}, we design the camera pose refinement module. 
By alternately optimizing 3D Gaussian primitives and camera poses during training, we improve the rendering quality of 3DGS and its robustness in dynamic scenes.

\paragraph{Iterative training.} 
Since the supervision of 3DGS primarily relies on photometric consistency loss, simultaneously optimizing camera parameters and 3DGS can be considered a chicken-and-egg problem. 
Therefore, similar to Bundle Adjustment, we adopt an alternating optimization strategy to train the model. 
Specifically, assuming \( G_t \) is the Gaussian at time \( t \), we first predict the deformation of the Gaussian using the deformation field \( \mathcal{D} \). We denote the deformed Gaussian as \( G_t^{t+1} \). Since the observation viewpoint changes from time \( t \) to \( t+1 \), \( G_t^{t+1} \) needs to be transformed once again under the camera \( C_{t+1} \) to render frame $I_{t+1}$. 
We denote the transformed Gaussian as \( G_{t+1} \).
This transformation process actually corresponds to camera motion. To achieve differentiable optimization, we introduce a small residual \( \Delta T \) into the relative pose \( T \) from camera viewpoint \( C_t \) to \( C_{t+1} \), treating it as a learnable SE(3) transformation. 
With this small change, we enable gradients to backpropagate to the camera poses. 
During the optimization of camera poses, we freeze all attributes of 3D Gaussians to improve training stability and robustness. 
Then we update the camera poses initialized by COLMAP with the optimized relative camera poses, achieving global pose refinement.

\paragraph{Discussion.}
While several methods have been proposed for pose-free optimization in static scenes, dynamic scenes present greater challenges due to their inherently under-constrained nature. As a result, to ensure stable and robust optimization, our approach still leverages camera poses computed by COLMAP as an initialization step. This also necessitates the presence of sufficient static features in the scene. Fortunately, static features are commonly found in most real-world environments, particularly in background regions.

\subsection{Optimization}
\label{sec:method_training}

Thanks to the integration of optical flow rendering and camera pose gradient computation in our rasterization process, the overall training pipeline of our method is end-to-end differentiable.
The overall training loss is given by:
\begin{equation}
    \mathcal{L} = \mathcal{L}_\text{baseline} + \lambda \mathcal{L}_\text{flow} ,
\end{equation}
where $\mathcal{L}_\text{baseline}$ is the photometric loss used in our baseline~\cite{yang2023deformable}, $\lambda$ is the weight of our flow loss. 

\section{Experiment}
\label{sec:Experiment}

\subsection{Experimental Setup}
\label{exp: setup}

To highlight the abilities of our method in handling complex dynamic scenes, we select two representative monocular dynamic scene datasets for evaluation: NeRF-DS~\cite{yan2023nerf} and HyperNeRF~\cite{park2021hypernerf}. 
Our implementation is mainly based on PyTorch. 
We use a simple Adam~\cite{kingma2014adam} optimizer to adjust the rotation increment and translation increment of the camera, and the learning rates of the two are set to 3e-3 and 1e-1, respectively. 
The entire training process requires 20,000 iterations.
We set $\lambda$ to 0.5 for NeRF-DS and 0.1 for HyperNeRF scene. 
The rest of the settings are consistent with the baseline method~\cite{yang2023deformable}. 
All experiments are performed on a single Nvidia RTX 3090 GPU.
For more implementation details, please refer to~\Cref{appendix: details}.

\subsection{Results}

Following previous methods, we use metrics PSNR, SSIM, and LPIPS for evaluation. 
For more visualizations, please refer to~\Cref{appendix: visualizations}.

\paragraph{Results on the NeRF-DS dataset.}

\Cref{tab: nerfds-per} shows the performance comparison results with the state-of-the-art methods on the NeRF-DS dataset. 
In dynamic monocular scenes, especially in those with rapid movements and high complexity, our method significantly outperforms the baseline method. 
For example (see~\Cref{fig: nerfds,fig: nerfds_full}), in the plate scene, our method accurately renders the reflections and sharp edges of the moving plate while significantly reducing visual distortions such as floating artifacts. Similarly, in the basin scene, our method effectively models the smooth surface of the basin, in contrast to other methods that result in a bumpy basin bottom.
This is mainly because our proposed framework can provide accurate and effective motion guidance for Gaussian deformation. 

\paragraph{Results on the HyperNeRF dataset.}
For scenes captured in the wild using smartphones, 
\Cref{tab: hypernerf_comparison} summarizes the relevant performance comparison results. Our method also achieves consistent performance improvements in these scenarios. 
Qualitatively, our approach excels at accurately reconstructing scene geometry and appearance, even under irregular camera movements and inaccurate camera poses. For instance (see~\Cref{fig: hypernerf,fig: hyper_full}), in the chicken scene, our method captures the subtle bumps on the red shell, while in the broom scene, it accurately renders the details of the broom in fast movement.
This is mainly attributed to the motion guidance and camera pose refinement proposed by our method, both of which enhance the rendering performance of the baseline method.

\begin{table}[t]
\centering
\caption{\textbf{Quantitative comparison on NeRF-DS dataset per-scene}. We highlight the \colorbox{pink}{best} and the \colorbox{yellow}{second best} results in each scene. NeRF-DS and HyperNeRF employ MS-SSIM and LPIPS with the AlexNet, while other methods and ours use SSIM and LPIPS with the VGG network.}
\resizebox{\textwidth}{!}{
\begin{tabular}{ccccccccccccccccc}
\toprule 
\multirow[c]{2}{*}{ Method } & \multicolumn{3}{c}{ Sieve } & \multicolumn{3}{c}{ Plate } & \multicolumn{3}{c}{ Bell } & \multicolumn{3}{c}{ Press } \\
\cmidrule(lr){2-4}\cmidrule(lr){5-7}\cmidrule(lr){8-10}\cmidrule(lr){11-13}
& PSNR$\uparrow$ & SSIM$\uparrow$ & LPIPS$\downarrow$ & PSNR$\uparrow$ & SSIM$\uparrow$ & LPIPS$\downarrow$ & PSNR$\uparrow$ & SSIM$\uparrow$ & LPIPS$\downarrow$ & PSNR$\uparrow$ & SSIM $\uparrow$ & LPIPS$\downarrow$ \\
\midrule 
3D-GS~\cite{kerbl20233d}                  & 23.16          & 0.8203         & 0.2247            & 16.14          & 0.6970         & 0.4093            & 21.01          & 0.7885         & 0.2503            & 22.89          & 0.8163          & 0.2904            \\
TiNeuVox~\cite{fang2022fast}              & 21.49          & 0.8265         & 0.3176            & \cellcolor{yellow}20.58          & 0.8027         & 0.3317            & 23.08          & 0.8242         & 0.2568            & 24.47          & 0.8613          & 0.3001            \\
HyperNeRF~\cite{park2021hypernerf}        & 25.43          & 0.8798         & 0.1645            & 18.93          & 0.7709         & 0.2940            & 23.06          & 0.8097         & 0.2052            & \cellcolor{yellow}26.15          & \cellcolor{pink}0.8897          & 0.1959            \\
NeRF-DS~\cite{yan2023nerf}                & \cellcolor{yellow}25.78          & \cellcolor{pink}0.8900         & \cellcolor{pink}0.1472            & 20.54          & \cellcolor{yellow}0.8042         & \cellcolor{yellow}0.1996            & 23.19          & 0.8212         & 0.1867            & 25.72          & 0.8618          & 0.2047            \\
Deformable-3DGS~\cite{yang2023deformable} & 25.14          & 0.8674         & \cellcolor{yellow}0.1502            & 18.82          & 0.7404         & 0.3554            & \cellcolor{yellow}25.42          & \cellcolor{yellow}0.8481         & \cellcolor{yellow}0.1570            & 25.41          & 0.8614          & \cellcolor{pink}0.1918            \\
Ours                                                            & \cellcolor{pink}26.17          & \cellcolor{yellow}0.8884         & \cellcolor{yellow}0.1502            & \cellcolor{pink}21.01          & \cellcolor{pink}0.8213         & \cellcolor{pink}0.1907            & \cellcolor{pink}26.33          & \cellcolor{pink}0.8688         & \cellcolor{pink}0.1490            & \cellcolor{pink}26.63          & \cellcolor{yellow}0.8865          & \cellcolor{yellow}0.1955            \\ 
\midrule
 \multirow{2}{*}{Method} & \multicolumn{3}{c}{Cup}                             & \multicolumn{3}{c}{As}                              & \multicolumn{3}{c}{Basin}                           & \multicolumn{3}{c}{Mean}                             \\
\cmidrule(lr){2-4}\cmidrule(lr){5-7}\cmidrule(lr){8-10}\cmidrule(lr){11-13}
 & PSNR$\uparrow$ & SSIM$\uparrow$ & LPIPS$\downarrow$ & PSNR$\uparrow$ & SSIM$\uparrow$ & LPIPS$\downarrow$ & PSNR$\uparrow$ & SSIM$\uparrow$ & LPIPS$\downarrow$ & PSNR$\uparrow$ & SSIM $\uparrow$ & LPIPS$\downarrow$ \\ 
\midrule
3D-GS~\cite{kerbl20233d}                  & 21.71          & 0.8304         & 0.2548            & 22.69          & 0.8017         & 0.2994            & 18.42          & 0.7170         & 0.3153            & 20.29          & 0.7816          & 0.2920            \\
TiNeuVox~\cite{fang2022fast}              & 19.71          & 0.8109         & 0.3643            & 21.26          & 0.8289         & 0.3967            & \cellcolor{pink}20.66          & 0.8145         & 0.2690            & 21.61          & 0.8234          & 0.2766            \\
HyperNeRF~\cite{park2021hypernerf}        & 24.59          & 0.8770         & 0.1650            & 25.58          & \cellcolor{pink}0.8949         & 0.1777            & \cellcolor{yellow}20.41          & \cellcolor{pink}0.8199         & 0.1911            & 23.45          & 0.8488          & 0.1990            \\
NeRF-DS~\cite{yan2023nerf}                & \cellcolor{yellow}24.91          & 0.8741         & 0.1737            & 25.13          & 0.8778         & \cellcolor{pink}0.1741            & 19.96          & \cellcolor{yellow}0.8166         & \cellcolor{pink}0.1855            & 23.60          & \cellcolor{yellow}0.8494          & \cellcolor{yellow}0.1816            \\
Deformable-3DGS~\cite{yang2023deformable} & 24.76          & \cellcolor{yellow}0.8876         & \cellcolor{pink}0.1544            & \cellcolor{yellow}26.08          & 0.8827         & 0.1832            & 19.61          & 0.7888         & 0.1871            & \cellcolor{yellow}23.61          & 0.8394          & 0.1970            \\
Ours                                                            & \cellcolor{pink}24.97          & \cellcolor{pink}0.8916         & \cellcolor{yellow}0.1556            & \cellcolor{pink}26.56          & \cellcolor{yellow}0.8902         & \cellcolor{yellow}0.1757            & 20.11          & 0.8126         & \cellcolor{yellow}0.1865            & \cellcolor{pink}24.54          & \cellcolor{pink}0.8656          & \cellcolor{pink}0.1719            \\ 
\bottomrule
\end{tabular}}
\label{tab: nerfds-per}
\end{table}

\begin{table}[ht]
\centering  
\caption{\textbf{Quantitative comparison on HyperNeRF's vrig dataset per-scene.}}  
\label{tab: hypernerf_comparison} 
\resizebox{\linewidth}{!}{
\begin{tabular}{c cc cc cc cc cc}  
\toprule  
\multirow{2}{*}{Method}&  
\multicolumn{2}{c}{3D Printer}&\multicolumn{2}{c}{Chicken}&\multicolumn{2}{c}{Broom}&\multicolumn{2}{c}{Banana}&\multicolumn{2}{c}{Mean}\cr  
\cmidrule(lr){2-3}\cmidrule(lr){4-5}\cmidrule(lr){6-7}\cmidrule(lr){8-9}\cmidrule(lr){10-11}
&PSNR$\uparrow$&SSIM$\uparrow$&PSNR$\uparrow$&SSIM$\uparrow$&PSNR$\uparrow$&SSIM$\uparrow$&PSNR$\uparrow$&SSIM$\uparrow$&PSNR$\uparrow$&SSIM$\uparrow$\cr  
\midrule
HyperNeRF~\cite{park2021hypernerf}&20.0&0.63&27.4&0.63&19.5&0.21&22.1&0.72&22.3&0.55\cr
TiNeuVox~\cite{fang2022fast}&\cellcolor{pink}22.8&\cellcolor{pink}0.73&\cellcolor{pink}28.2&\cellcolor{pink}0.79&\cellcolor{yellow}21.3&0.31&24.4&0.64&\cellcolor{yellow}24.2&\cellcolor{yellow}0.62\cr  
Deformable 3DGS~\cite{yang2023deformable}&20.5&0.64&22.8&0.61&20.5&\cellcolor{yellow}0.35&\cellcolor{yellow}26.0&\cellcolor{yellow}0.83& 22.5 & 0.61\cr
Ours& \cellcolor{yellow}21.8 & \cellcolor{yellow}0.71 & \cellcolor{yellow}26.8 & \cellcolor{pink}0.79 & \cellcolor{pink}22.3 & \cellcolor{pink}0.38 & \cellcolor{pink}28.2 & \cellcolor{pink}0.86 & \cellcolor{pink}24.8 & \cellcolor{pink}0.69 \cr
\bottomrule  
\end{tabular}  
}
\end{table}

\subsection{Ablation Study}
\label{exp: ablation study}
In this section, we conduct ablations on the NeRF-DS dataset to validate the effectiveness of the key components of our method, as shown in~\Cref{tab:flow-ablation}.
For more ablations, please refer to~\Cref{appendix: ablations}.

\begin{table}[htbp]
\centering
\caption{\textbf{Ablations on the key components of our proposed framework.}
}
\label{tab:flow-ablation}%
\setlength{\tabcolsep}{1em}{
\begin{tabular}{l|ccc}
    \toprule
    Setting & PSNR $\uparrow$ & SSIM $\uparrow$ & LPIPS $\downarrow$ \\
    \midrule
    Baseline & 23.61  & 0.8394  & 0.1970  \\
    + optical flow guidance & 23.37  & 0.8333  & 0.2112  \\
    + motion flow guidance  & \cellcolor{yellow}24.12 & \cellcolor{yellow}0.8609 & \cellcolor{yellow}0.1763 \\
    + motion flow guidance + camera pose refinement & \cellcolor{pink}24.54 & \cellcolor{pink}0.8656 & \cellcolor{pink}0.1719 \\    
    \bottomrule
\end{tabular}%
}
\end{table}%

\begin{figure}[t]
   \centering
   \includegraphics[width=\linewidth]{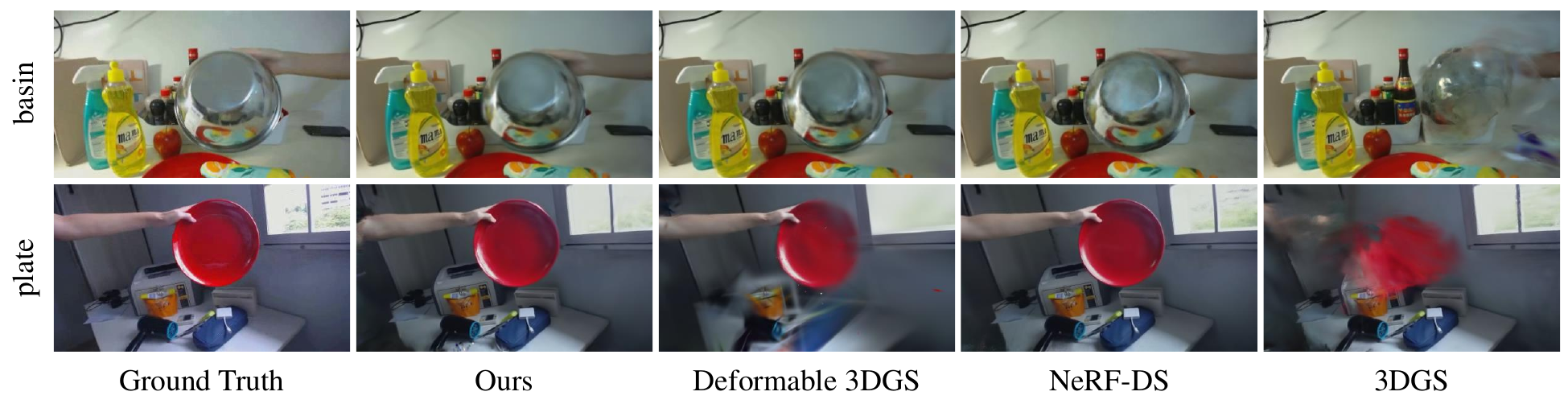}
   \vspace{-1.5em}
   \caption{\textbf{Qualitative comparison on NeRF-DS dataset.} Refer to~\Cref{fig: nerfds_full} for more scenes.}
   \label{fig: nerfds}
\vspace{1.5em}
   \centering
   \includegraphics[width=\linewidth]{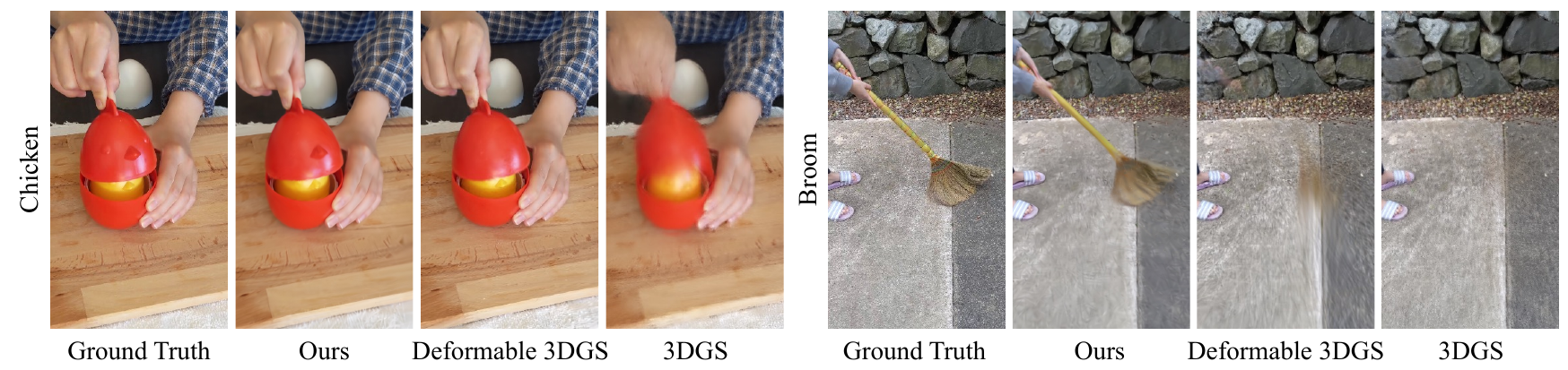}
   \vspace{-1.5em}
   \caption{\textbf{Qualitative comparison on HyperNeRF dataset.} Refer to~\Cref{fig: hyper_full} for more scenes.}
   \label{fig: hypernerf}
\vspace{1.5em}
   \centering
   \includegraphics[width=0.95\linewidth]{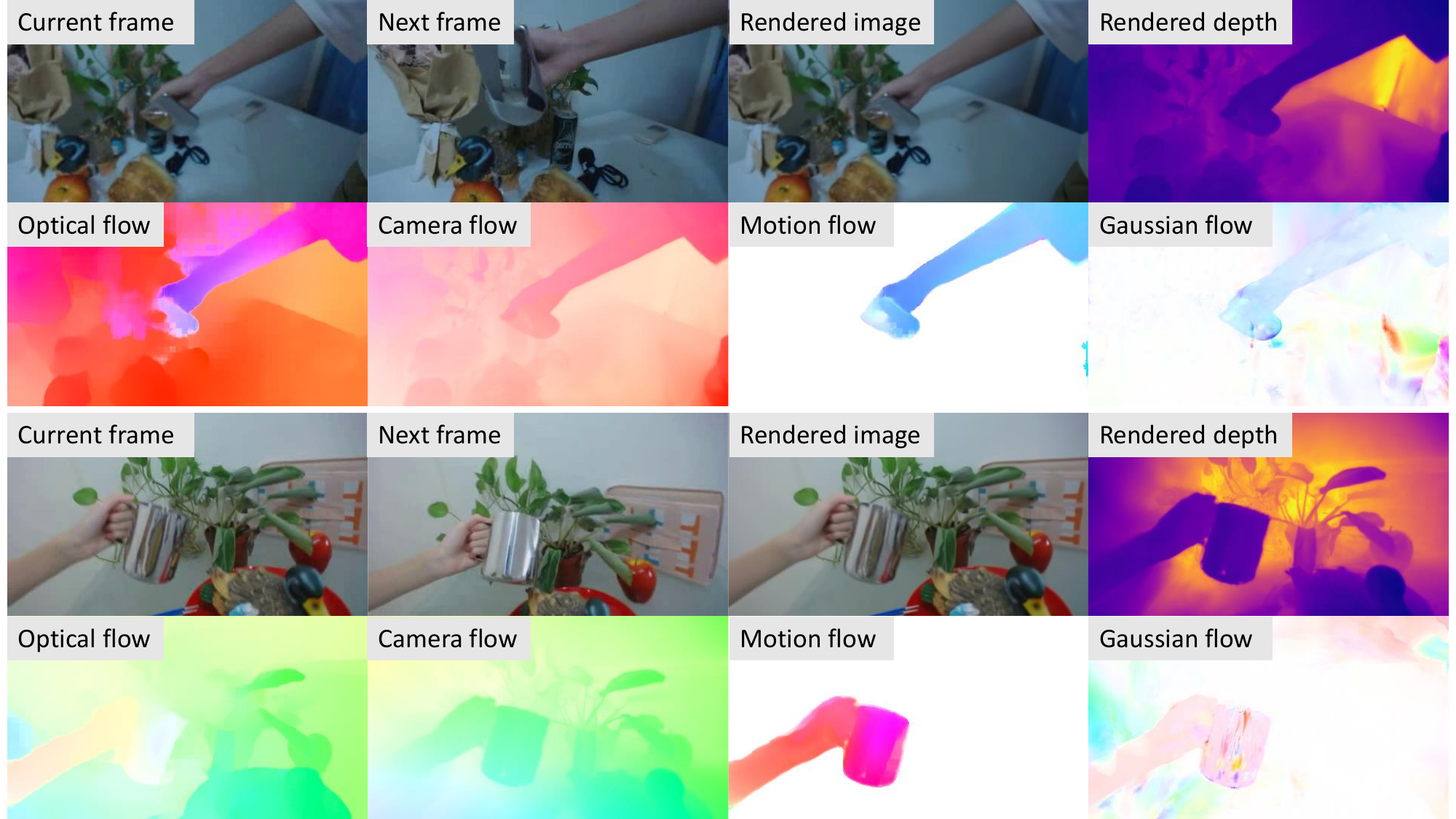}
   \caption{\textbf{Visualization of all data flows.} 
   Each example corresponds to two rows. 
   }
   \label{fig: example}
\end{figure}

\paragraph{Effectiveness of the optical flow decoupling module.}
To illustrate the necessity of optical flow decoupling module, we conduct ablations using direct optical flow supervision instead of decoupled motion flow constraints. 
As shown in the row 2 of~\Cref{tab:flow-ablation}, directly using optical flow to supervise Gaussian motion even results in a performance decline compared to the baseline. This performance drop is likely due to the inherent ambiguity created by the mixed camera and object movements. When camera and object movements are not separated, the supervision signal becomes noisy and less effective. This ambiguity hampers the optimization process of 3DGS, thereby reducing the motion modeling capabilities of the deformation field. 
In contrast, we use motion flow as supervision, which effectively provides explicit motion guidance for Gaussian deformation, thereby better modeling complex dynamic scenes.
As shown in~\Cref{fig: example,,fig: example_full}, only the motion flow can clearly highlights movement information in dynamic regions. This explicit motion information efficiently constrains the Gaussian flow, ensuring that the motion guidance remains consistent and effective.

\begin{figure}[t]
   \centering
   \includegraphics[width=\linewidth]{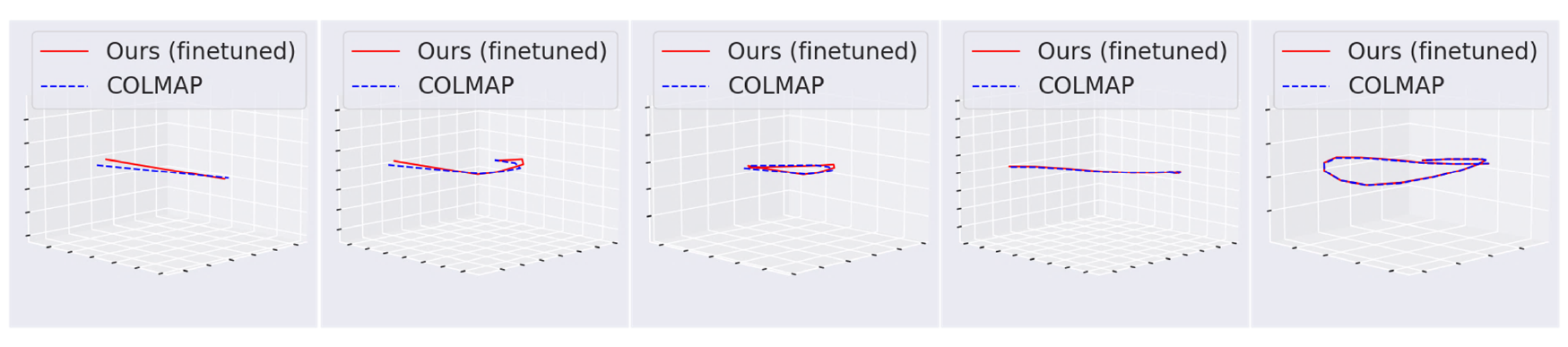}
   \caption{\textbf{Visualization of the camera trajectories optimized by our method and COLMAP.} 
   }
   \label{fig: pose}
\end{figure}

\paragraph{Effectiveness of the camera pose refinement module.}
Leveraging the alternating optimization of 3DGS and camera poses, our approach adaptively corrects potential errors in camera poses. Furthermore, the updated camera poses contribute to more accurate camera flow, thus improving the accuracy of motion guidance. In row 4 of~\Cref{tab:flow-ablation}, our camera pose refinement module, built upon motion guidance, yields substantial performance gains for the model. This iterative optimization process enhances the robustness of our model in complex dynamic scenes. For instance, in the HyperNeRF dataset, our method reconstructs more plausible results compared to the baseline approach. Unlike static scene datasets (e.g., Tanks \& Temples) that use COLMAP to obtain the ground truth of camera poses, we assume that COLMAP may not provide accurate poses for dynamic scene datasets. In this setting, we lack ground truth for a direct quantitative evaluation for refined camera pose. Therefore, we provide visualizations of the pose refinement process in the~\Cref{fig: pose} as qualitative comparison. 

\section{Conclusion}
\label{sec:Conclusion}

In this paper, we propose MotionGS, a novel deformable 3D Gaussian Splatting framework for explicitly modeling and constraining object motion in dynamic scene reconstruction. The proposed framework includes two key modules: the optical flow decoupling module and the camera poserefinement module. 
The optical flow decoupling module decouples the motion flow related solely to object motion from the optical flow priors, providing explicit supervision for Gaussian deformation.
The camera pose refinement module alternately optimizes 3DGS and camera poses, further enhancing the rendering quality and robustness of our model in dynamic scenes. 
Quantitative and qualitative results on the NeRF-DS and HyperNeRF datasets strongly demonstrate the contributions and effectiveness of our proposed method. 
More importantly, the proposed improvements are agnostic to specific network designs, which can be applied to similar deformation-based 3DGS methods. 
In future work, we aim to develop a 3DGS method that does not rely on camera pose inputs, thereby achieving robust high-quality reconstruction in dynamic scenes.

{
\small
\bibliographystyle{unsrt}
\bibliography{ref}

\begin{thebibliography}{10}

\bibitem{pumarola2021d}
Albert Pumarola, Enric Corona, Gerard Pons-Moll, and Francesc Moreno-Noguer.
\newblock D-nerf: Neural radiance fields for dynamic scenes.
\newblock In {\em Proceedings of the IEEE/CVF Conference on Computer Vision and Pattern Recognition}, pages 10318--10327, 2021.

\bibitem{park2021nerfies}
Keunhong Park, Utkarsh Sinha, Jonathan~T Barron, Sofien Bouaziz, Dan~B Goldman, Steven~M Seitz, and Ricardo Martin-Brualla.
\newblock Nerfies: Deformable neural radiance fields.
\newblock In {\em Proceedings of the IEEE/CVF International Conference on Computer Vision}, pages 5865--5874, 2021.

\bibitem{li2021neural}
Zhengqi Li, Simon Niklaus, Noah Snavely, and Oliver Wang.
\newblock Neural scene flow fields for space-time view synthesis of dynamic scenes.
\newblock In {\em Proceedings of the IEEE/CVF Conference on Computer Vision and Pattern Recognition}, pages 6498--6508, 2021.

\bibitem{liu2023robust}
Yu-Lun Liu, Chen Gao, Andreas Meuleman, Hung-Yu Tseng, Ayush Saraf, Changil Kim, Yung-Yu Chuang, Johannes Kopf, and Jia-Bin Huang.
\newblock Robust dynamic radiance fields.
\newblock In {\em Proceedings of the IEEE/CVF Conference on Computer Vision and Pattern Recognition}, pages 13--23, 2023.

\bibitem{gao2022monocular}
Hang Gao, Ruilong Li, Shubham Tulsiani, Bryan Russell, and Angjoo Kanazawa.
\newblock Monocular dynamic view synthesis: A reality check.
\newblock {\em Advances in Neural Information Processing Systems}, 35:33768--33780, 2022.

\bibitem{li2023dynibar}
Zhengqi Li, Qianqian Wang, Forrester Cole, Richard Tucker, and Noah Snavely.
\newblock Dynibar: Neural dynamic image-based rendering.
\newblock In {\em Proceedings of the IEEE/CVF Conference on Computer Vision and Pattern Recognition}, pages 4273--4284, 2023.

\bibitem{cao2023hexplane}
Ang Cao and Justin Johnson.
\newblock Hexplane: A fast representation for dynamic scenes.
\newblock In {\em Proceedings of the IEEE/CVF Conference on Computer Vision and Pattern Recognition}, pages 130--141, 2023.

\bibitem{shao2023tensor4d}
Ruizhi Shao, Zerong Zheng, Hanzhang Tu, Boning Liu, Hongwen Zhang, and Yebin Liu.
\newblock Tensor4d: Efficient neural 4d decomposition for high-fidelity dynamic reconstruction and rendering.
\newblock In {\em Proceedings of the IEEE/CVF Conference on Computer Vision and Pattern Recognition}, pages 16632--16642, 2023.

\bibitem{mildenhall2021nerf}
Ben Mildenhall, Pratul~P Srinivasan, Matthew Tancik, Jonathan~T Barron, Ravi Ramamoorthi, and Ren Ng.
\newblock Nerf: Representing scenes as neural radiance fields for view synthesis.
\newblock {\em Communications of the ACM}, 65(1):99--106, 2021.

\bibitem{kerbl20233d}
Bernhard Kerbl, Georgios Kopanas, Thomas Leimk{\"u}hler, and George Drettakis.
\newblock 3d gaussian splatting for real-time radiance field rendering.
\newblock {\em ACM Transactions on Graphics}, 42(4):1--14, 2023.

\bibitem{luiten2023dynamic}
Jonathon Luiten, Georgios Kopanas, Bastian Leibe, and Deva Ramanan.
\newblock Dynamic 3d gaussians: Tracking by persistent dynamic view synthesis.
\newblock {\em arXiv preprint arXiv:2308.09713}, 2023.

\bibitem{li2023spacetime}
Zhan Li, Zhang Chen, Zhong Li, and Yi~Xu.
\newblock Spacetime gaussian feature splatting for real-time dynamic view synthesis.
\newblock {\em arXiv preprint arXiv:2312.16812}, 2023.

\bibitem{wu20234d}
Guanjun Wu, Taoran Yi, Jiemin Fang, Lingxi Xie, Xiaopeng Zhang, Wei Wei, Wenyu Liu, Qi~Tian, and Xinggang Wang.
\newblock 4d gaussian splatting for real-time dynamic scene rendering.
\newblock {\em arXiv preprint arXiv:2310.08528}, 2023.

\bibitem{yang2023gs4d}
Zeyu Yang, Hongye Yang, Zijie Pan, and Li~Zhang.
\newblock Real-time photorealistic dynamic scene representation and rendering with 4d gaussian splatting.
\newblock {\em International Conference on Learning Representations (ICLR)}, 2024.

\bibitem{huang2023sc}
Yi-Hua Huang, Yang-Tian Sun, Ziyi Yang, Xiaoyang Lyu, Yan-Pei Cao, and Xiaojuan Qi.
\newblock Sc-gs: Sparse-controlled gaussian splatting for editable dynamic scenes.
\newblock {\em arXiv preprint arXiv:2312.14937}, 2023.

\bibitem{lin2023gaussian}
Youtian Lin, Zuozhuo Dai, Siyu Zhu, and Yao Yao.
\newblock Gaussian-flow: 4d reconstruction with dynamic 3d gaussian particle.
\newblock {\em arXiv preprint arXiv:2312.03431}, 2023.

\bibitem{yang2023deformable}
Ziyi Yang, Xinyu Gao, Wen Zhou, Shaohui Jiao, Yuqing Zhang, and Xiaogang Jin.
\newblock Deformable 3d gaussians for high-fidelity monocular dynamic scene reconstruction.
\newblock {\em arXiv preprint arXiv:2309.13101}, 2023.

\bibitem{duan20244d}
Yuanxing Duan, Fangyin Wei, Qiyu Dai, Yuhang He, Wenzheng Chen, and Baoquan Chen.
\newblock 4d gaussian splatting: Towards efficient novel view synthesis for dynamic scenes.
\newblock {\em arXiv preprint arXiv:2402.03307}, 2024.

\bibitem{huang2022flowformer}
Zhaoyang Huang, Xiaoyu Shi, Chao Zhang, Qiang Wang, Ka~Chun Cheung, Hongwei Qin, Jifeng Dai, and Hongsheng Li.
\newblock Flowformer: A transformer architecture for optical flow.
\newblock In {\em European conference on computer vision}, pages 668--685. Springer, 2022.

\bibitem{xu2022gmflow}
Haofei Xu, Jing Zhang, Jianfei Cai, Hamid Rezatofighi, and Dacheng Tao.
\newblock Gmflow: Learning optical flow via global matching.
\newblock In {\em Proceedings of the IEEE/CVF Conference on Computer Vision and Pattern Recognition}, pages 8121--8130, 2022.

\bibitem{gao2024gaussianflow}
Quankai Gao, Qiangeng Xu, Zhe Cao, Ben Mildenhall, Wenchao Ma, Le~Chen, Danhang Tang, and Ulrich Neumann.
\newblock Gaussianflow: Splatting gaussian dynamics for 4d content creation.
\newblock {\em arXiv preprint arXiv:2403.12365}, 2024.

\bibitem{guo2024motion}
Zhiyang Guo, Wengang Zhou, Li~Li, Min Wang, and Houqiang Li.
\newblock Motion-aware 3d gaussian splatting for efficient dynamic scene reconstruction.
\newblock {\em arXiv preprint arXiv:2403.11447}, 2024.

\bibitem{barron2021mip}
Jonathan~T Barron, Ben Mildenhall, Matthew Tancik, Peter Hedman, Ricardo Martin-Brualla, and Pratul~P Srinivasan.
\newblock Mip-nerf: A multiscale representation for anti-aliasing neural radiance fields.
\newblock In {\em Proceedings of the IEEE/CVF International Conference on Computer Vision}, pages 5855--5864, 2021.

\bibitem{barron2022mip}
Jonathan~T Barron, Ben Mildenhall, Dor Verbin, Pratul~P Srinivasan, and Peter Hedman.
\newblock Mip-nerf 360: Unbounded anti-aliased neural radiance fields.
\newblock In {\em Proceedings of the IEEE/CVF Conference on Computer Vision and Pattern Recognition}, pages 5470--5479, 2022.

\bibitem{sun2022direct}
Cheng Sun, Min Sun, and Hwann-Tzong Chen.
\newblock Direct voxel grid optimization: Super-fast convergence for radiance fields reconstruction.
\newblock In {\em Proceedings of the IEEE/CVF Conference on Computer Vision and Pattern Recognition}, pages 5459--5469, 2022.

\bibitem{xu2022point}
Qiangeng Xu, Zexiang Xu, Julien Philip, Sai Bi, Zhixin Shu, Kalyan Sunkavalli, and Ulrich Neumann.
\newblock Point-nerf: Point-based neural radiance fields.
\newblock In {\em Proceedings of the IEEE/CVF conference on computer vision and pattern recognition}, pages 5438--5448, 2022.

\bibitem{wang2021ibrnet}
Qianqian Wang, Zhicheng Wang, Kyle Genova, Pratul~P Srinivasan, Howard Zhou, Jonathan~T Barron, Ricardo Martin-Brualla, Noah Snavely, and Thomas Funkhouser.
\newblock Ibrnet: Learning multi-view image-based rendering.
\newblock In {\em Proceedings of the IEEE/CVF Conference on Computer Vision and Pattern Recognition}, pages 4690--4699, 2021.

\bibitem{muller2022instant}
Thomas M{\"u}ller, Alex Evans, Christoph Schied, and Alexander Keller.
\newblock Instant neural graphics primitives with a multiresolution hash encoding.
\newblock {\em ACM transactions on graphics (TOG)}, 41(4):1--15, 2022.

\bibitem{yu2021plenoctrees}
Alex Yu, Ruilong Li, Matthew Tancik, Hao Li, Ren Ng, and Angjoo Kanazawa.
\newblock Plenoctrees for real-time rendering of neural radiance fields.
\newblock In {\em Proceedings of the IEEE/CVF International Conference on Computer Vision}, pages 5752--5761, 2021.

\bibitem{zhang2020nerf++}
Kai Zhang, Gernot Riegler, Noah Snavely, and Vladlen Koltun.
\newblock Nerf++: Analyzing and improving neural radiance fields.
\newblock {\em arXiv preprint arXiv:2010.07492}, 2020.

\bibitem{zhu2023tiface}
Ruijie Zhu, Jiahao Chang, Ziyang Song, Jiahuan Yu, and Tianzhu Zhang.
\newblock Tiface: Improving facial reconstruction through tensorial radiance fields and implicit surfaces.
\newblock {\em arXiv preprint arXiv:2312.09527}, 2023.

\bibitem{li2024dngaussian}
Jiahe Li, Jiawei Zhang, Xiao Bai, Jin Zheng, Xin Ning, Jun Zhou, and Lin Gu.
\newblock Dngaussian: Optimizing sparse-view 3d gaussian radiance fields with global-local depth normalization.
\newblock {\em arXiv preprint arXiv:2403.06912}, 2024.

\bibitem{chen2024mvsplat}
Yuedong Chen, Haofei Xu, Chuanxia Zheng, Bohan Zhuang, Marc Pollefeys, Andreas Geiger, Tat-Jen Cham, and Jianfei Cai.
\newblock Mvsplat: Efficient 3d gaussian splatting from sparse multi-view images.
\newblock {\em arXiv preprint arXiv:2403.14627}, 2024.

\bibitem{charatan2024pixelsplat}
David Charatan, Sizhe Li, Andrea Tagliasacchi, and Vincent Sitzmann.
\newblock pixelsplat: 3d gaussian splats from image pairs for scalable generalizable 3d reconstruction, 2024.

\bibitem{szymanowicz2024splatter_image}
Stanislaw Szymanowicz, Christian Rupprecht, and Andrea Vedaldi.
\newblock Splatter image: Ultra-fast single-view 3d reconstruction.
\newblock {\em Conference on Computer Vision and Pattern Recognition (CVPR)}, 2024.

\bibitem{tang2023dreamgaussian}
Jiaxiang Tang, Jiawei Ren, Hang Zhou, Ziwei Liu, and Gang Zeng.
\newblock Dreamgaussian: Generative gaussian splatting for efficient 3d content creation.
\newblock {\em arXiv preprint arXiv:2309.16653}, 2023.

\bibitem{blattmann2023align}
Andreas Blattmann, Robin Rombach, Huan Ling, Tim Dockhorn, Seung~Wook Kim, Sanja Fidler, and Karsten Kreis.
\newblock Align your latents: High-resolution video synthesis with latent diffusion models.
\newblock In {\em Proceedings of the IEEE/CVF Conference on Computer Vision and Pattern Recognition}, pages 22563--22575, 2023.

\bibitem{tang2024lgm}
Jiaxiang Tang, Zhaoxi Chen, Xiaokang Chen, Tengfei Wang, Gang Zeng, and Ziwei Liu.
\newblock Lgm: Large multi-view gaussian model for high-resolution 3d content creation.
\newblock {\em arXiv preprint arXiv:2402.05054}, 2024.

\bibitem{wang2023prolificdreamer}
Zhengyi Wang, Cheng Lu, Yikai Wang, Fan Bao, Chongxuan Li, Hang Su, and Jun Zhu.
\newblock Prolificdreamer: High-fidelity and diverse text-to-3d generation with variational score distillation.
\newblock {\em arXiv preprint arXiv:2305.16213}, 2023.

\bibitem{xu2024grm}
Xu~Yinghao, Shi Zifan, Yifan Wang, Chen Hansheng, Yang Ceyuan, Peng Sida, Shen Yujun, and Wetzstein Gordon.
\newblock Grm: Large gaussian reconstruction model for efficient 3d reconstruction and generation, 2024.

\bibitem{chen2023gaussianeditor}
Yiwen Chen, Zilong Chen, Chi Zhang, Feng Wang, Xiaofeng Yang, Yikai Wang, Zhongang Cai, Lei Yang, Huaping Liu, and Guosheng Lin.
\newblock Gaussianeditor: Swift and controllable 3d editing with gaussian splatting, 2023.

\bibitem{GaussianEditor}
Jiemin Fang, Junjie Wang, Xiaopeng Zhang, Lingxi Xie, and Qi~Tian.
\newblock Gaussianeditor: Editing 3d gaussians delicately with text instructions.
\newblock In {\em CVPR}, 2024.

\bibitem{zhou2023feature}
Shijie Zhou, Haoran Chang, Sicheng Jiang, Zhiwen Fan, Zehao Zhu, Dejia Xu, Pradyumna Chari, Suya You, Zhangyang Wang, and Achuta Kadambi.
\newblock Feature 3dgs: Supercharging 3d gaussian splatting to enable distilled feature fields.
\newblock {\em arXiv preprint arXiv:2312.03203}, 2023.

\bibitem{yan2023gs}
Chi Yan, Delin Qu, Dan Xu, Bin Zhao, Zhigang Wang, Dong Wang, and Xuelong Li.
\newblock Gs-slam: Dense visual slam with 3d gaussian splatting.
\newblock In {\em CVPR}, 2024.

\bibitem{keetha2024splatam}
Nikhil Keetha, Jay Karhade, Krishna~Murthy Jatavallabhula, Gengshan Yang, Sebastian Scherer, Deva Ramanan, and Jonathon Luiten.
\newblock Splatam: Splat, track \& map 3d gaussians for dense rgb-d slam.
\newblock In {\em Proceedings of the IEEE/CVF Conference on Computer Vision and Pattern Recognition}, 2024.

\bibitem{Matsuki:Murai:etal:CVPR2024}
Hidenobu Matsuki, Riku Murai, Paul H.~J. Kelly, and Andrew~J. Davison.
\newblock {G}aussian {S}platting {SLAM}.
\newblock In {\em Proceedings of the IEEE/CVF Conference on Computer Vision and Pattern Recognition}, 2024.

\bibitem{hhuang2024photoslam}
Huajian Huang, Longwei Li, Cheng Hui, and Sai-Kit Yeung.
\newblock Photo-slam: Real-time simultaneous localization and photorealistic mapping for monocular, stereo, and rgb-d cameras.
\newblock In {\em Proceedings of the IEEE/CVF Conference on Computer Vision and Pattern Recognition}, 2024.

\bibitem{du2021neural}
Yilun Du, Yinan Zhang, Hong-Xing Yu, Joshua~B Tenenbaum, and Jiajun Wu.
\newblock Neural radiance flow for 4d view synthesis and video processing.
\newblock In {\em 2021 IEEE/CVF International Conference on Computer Vision (ICCV)}, pages 14304--14314. IEEE Computer Society, 2021.

\bibitem{park2021hypernerf}
Keunhong Park, Utkarsh Sinha, Peter Hedman, Jonathan~T Barron, Sofien Bouaziz, Dan~B Goldman, Ricardo Martin-Brualla, and Steven~M Seitz.
\newblock Hypernerf: A higher-dimensional representation for topologically varying neural radiance fields.
\newblock {\em arXiv preprint arXiv:2106.13228}, 2021.

\bibitem{wang2023tracking}
Qianqian Wang, Yen-Yu Chang, Ruojin Cai, Zhengqi Li, Bharath Hariharan, Aleksander Holynski, and Noah Snavely.
\newblock Tracking everything everywhere all at once.
\newblock In {\em Proceedings of the IEEE/CVF International Conference on Computer Vision}, pages 19795--19806, 2023.

\bibitem{yang2022banmo}
Gengshan Yang, Minh Vo, Natalia Neverova, Deva Ramanan, Andrea Vedaldi, and Hanbyul Joo.
\newblock Banmo: Building animatable 3d neural models from many casual videos.
\newblock In {\em Proceedings of the IEEE/CVF Conference on Computer Vision and Pattern Recognition}, pages 2863--2873, 2022.

\bibitem{xian2021space}
Wenqi Xian, Jia-Bin Huang, Johannes Kopf, and Changil Kim.
\newblock Space-time neural irradiance fields for free-viewpoint video.
\newblock In {\em Proceedings of the IEEE/CVF Conference on Computer Vision and Pattern Recognition}, pages 9421--9431, 2021.

\bibitem{wang2021neural}
Chaoyang Wang, Ben Eckart, Simon Lucey, and Orazio Gallo.
\newblock Neural trajectory fields for dynamic novel view synthesis.
\newblock {\em arXiv preprint arXiv:2105.05994}, 2021.

\bibitem{gao2021dynamic}
Chen Gao, Ayush Saraf, Johannes Kopf, and Jia-Bin Huang.
\newblock Dynamic view synthesis from dynamic monocular video.
\newblock In {\em Proceedings of the IEEE/CVF International Conference on Computer Vision}, pages 5712--5721, 2021.

\bibitem{zhou2024dynpoint}
Kaichen Zhou, Jia-Xing Zhong, Sangyun Shin, Kai Lu, Yiyuan Yang, Andrew Markham, and Niki Trigoni.
\newblock Dynpoint: Dynamic neural point for view synthesis.
\newblock {\em Advances in Neural Information Processing Systems}, 36, 2023.

\bibitem{fridovich2023k}
Sara Fridovich-Keil, Giacomo Meanti, Frederik~Rahb{\ae}k Warburg, Benjamin Recht, and Angjoo Kanazawa.
\newblock K-planes: Explicit radiance fields in space, time, and appearance.
\newblock In {\em Proceedings of the IEEE/CVF Conference on Computer Vision and Pattern Recognition}, pages 12479--12488, 2023.

\bibitem{xu20234k4d}
Zhen Xu, Sida Peng, Haotong Lin, Guangzhao He, Jiaming Sun, Yujun Shen, Hujun Bao, and Xiaowei Zhou.
\newblock 4k4d: Real-time 4d view synthesis at 4k resolution.
\newblock {\em arXiv preprint arXiv:2310.11448}, 2023.

\bibitem{wang2023mixed}
Feng Wang, Sinan Tan, Xinghang Li, Zeyue Tian, Yafei Song, and Huaping Liu.
\newblock Mixed neural voxels for fast multi-view video synthesis.
\newblock In {\em Proceedings of the IEEE/CVF International Conference on Computer Vision}, pages 19706--19716, 2023.

\bibitem{fang2022fast}
Jiemin Fang, Taoran Yi, Xinggang Wang, Lingxi Xie, Xiaopeng Zhang, Wenyu Liu, Matthias Nie{\ss}ner, and Qi~Tian.
\newblock Fast dynamic radiance fields with time-aware neural voxels.
\newblock In {\em SIGGRAPH Asia 2022 Conference Papers}, pages 1--9, 2022.

\bibitem{katsumata2023efficient}
Kai Katsumata, Duc~Minh Vo, and Hideki Nakayama.
\newblock An efficient 3d gaussian representation for monocular/multi-view dynamic scenes.
\newblock {\em arXiv preprint arXiv:2311.12897}, 2023.

\bibitem{lu2024gagaussian}
Zhicheng Lu, Xiang Guo, Le~Hui, Tianrui Chen, Ming Yang, Xiao Tang, Feng Zhu, and Yuchao Dai.
\newblock 3d geometry-aware deformable gaussian splatting for dynamic view synthesis.
\newblock In {\em Proceedings of the IEEE/CVF Conference on Computer Vision and Pattern Recognition}, 2024.

\bibitem{das2023neural}
Devikalyan Das, Christopher Wewer, Raza Yunus, Eddy Ilg, and Jan~Eric Lenssen.
\newblock Neural parametric gaussians for monocular non-rigid object reconstruction.
\newblock {\em arXiv preprint arXiv:2312.01196}, 2023.

\bibitem{sun20243dgstream}
Jiakai Sun, Han Jiao, Guangyuan Li, Zhanjie Zhang, Lei Zhao, and Wei Xing.
\newblock 3dgstream: On-the-fly training of 3d gaussians for efficient streaming of photo-realistic free-viewpoint videos.
\newblock {\em arXiv preprint arXiv:2403.01444}, 2024.

\bibitem{yen2021inerf}
Lin Yen-Chen, Pete Florence, Jonathan~T Barron, Alberto Rodriguez, Phillip Isola, and Tsung-Yi Lin.
\newblock inerf: Inverting neural radiance fields for pose estimation.
\newblock In {\em 2021 IEEE/RSJ International Conference on Intelligent Robots and Systems (IROS)}, pages 1323--1330. IEEE, 2021.

\bibitem{wang2021nerf}
Zirui Wang, Shangzhe Wu, Weidi Xie, Min Chen, and Victor~Adrian Prisacariu.
\newblock Nerf--: Neural radiance fields without known camera parameters.
\newblock {\em arXiv preprint arXiv:2102.07064}, 2021.

\bibitem{xia2022sinerf}
Yitong Xia, Hao Tang, Radu Timofte, and Luc Van~Gool.
\newblock Sinerf: Sinusoidal neural radiance fields for joint pose estimation and scene reconstruction.
\newblock {\em arXiv preprint arXiv:2210.04553}, 2022.

\bibitem{lin2021barf}
Chen-Hsuan Lin, Wei-Chiu Ma, Antonio Torralba, and Simon Lucey.
\newblock Barf: Bundle-adjusting neural radiance fields.
\newblock In {\em Proceedings of the IEEE/CVF International Conference on Computer Vision}, pages 5741--5751, 2021.

\bibitem{chng2022garf}
Shin-Fang Chng, Sameera Ramasinghe, Jamie Sherrah, and Simon Lucey.
\newblock Garf: Gaussian activated radiance fields for high fidelity reconstruction and pose estimation.
\newblock {\em arXiv e-prints}, pages arXiv--2204, 2022.

\bibitem{bian2023nope}
Wenjing Bian, Zirui Wang, Kejie Li, Jia-Wang Bian, and Victor~Adrian Prisacariu.
\newblock Nope-nerf: Optimising neural radiance field with no pose prior.
\newblock In {\em Proceedings of the IEEE/CVF Conference on Computer Vision and Pattern Recognition}, pages 4160--4169, 2023.

\bibitem{fu2023colmap}
Yang Fu, Sifei Liu, Amey Kulkarni, Jan Kautz, Alexei~A Efros, and Xiaolong Wang.
\newblock Colmap-free 3d gaussian splatting.
\newblock {\em arXiv preprint arXiv:2312.07504}, 2023.

\bibitem{schoenberger2016sfm}
Johannes~Lutz Sch\"{o}nberger and Jan-Michael Frahm.
\newblock Structure-from-motion revisited.
\newblock In {\em Conference on Computer Vision and Pattern Recognition (CVPR)}, 2016.

\bibitem{MatsukiCVPR2024}
Hidenobu Matsuki, Riku Murai, Paul H.~J. Kelly, and Andrew~J. Davison.
\newblock {G}aussian {S}platting {SLAM}.
\newblock In {\em Proceedings of the IEEE/CVF Conference on Computer Vision and Pattern Recognition}, 2024.

\bibitem{yan2023nerf}
Zhiwen Yan, Chen Li, and Gim~Hee Lee.
\newblock Nerf-ds: Neural radiance fields for dynamic specular objects.
\newblock In {\em Proceedings of the IEEE/CVF Conference on Computer Vision and Pattern Recognition}, pages 8285--8295, 2023.

\bibitem{kingma2014adam}
Diederik~P Kingma.
\newblock Adam: A method for stochastic optimization.
\newblock {\em arXiv preprint arXiv:1412.6980}, 2014.

\bibitem{Ranftl2022}
Ren\'{e} Ranftl, Katrin Lasinger, David Hafner, Konrad Schindler, and Vladlen Koltun.
\newblock Towards robust monocular depth estimation: Mixing datasets for zero-shot cross-dataset transfer.
\newblock {\em IEEE Transactions on Pattern Analysis and Machine Intelligence}, 44(3), 2022.

\bibitem{kong2022mdflow}
Lingtong Kong and Jie Yang.
\newblock Mdflow: Unsupervised optical flow learning by reliable mutual knowledge distillation.
\newblock {\em IEEE Transactions on Circuits and Systems for Video Technology}, 33(2):677--688, 2022.

\bibitem{zhu2024scaledepth}
Ruijie Zhu, Chuxin Wang, Ziyang Song, Li~Liu, Tianzhu Zhang, and Yongdong Zhang.
\newblock Scaledepth: Decomposing metric depth estimation into scale prediction and relative depth estimation.
\newblock {\em arXiv preprint arXiv:2407.08187}, 2024.

\bibitem{li2022neural}
Tianye Li, Mira Slavcheva, Michael Zollhoefer, Simon Green, Christoph Lassner, Changil Kim, Tanner Schmidt, Steven Lovegrove, Michael Goesele, Richard Newcombe, et~al.
\newblock Neural 3d video synthesis from multi-view video.
\newblock In {\em Proceedings of the IEEE/CVF Conference on Computer Vision and Pattern Recognition}, pages 5521--5531, 2022.

\end{thebibliography}
}

\clearpage

\appendix

\section{Appendix}

\subsection{Formulation of Gaussian Flow}
\label{appendix: gaussian flow}

\begin{figure}[htbp]
   \centering
   \includegraphics[width=0.8\linewidth]{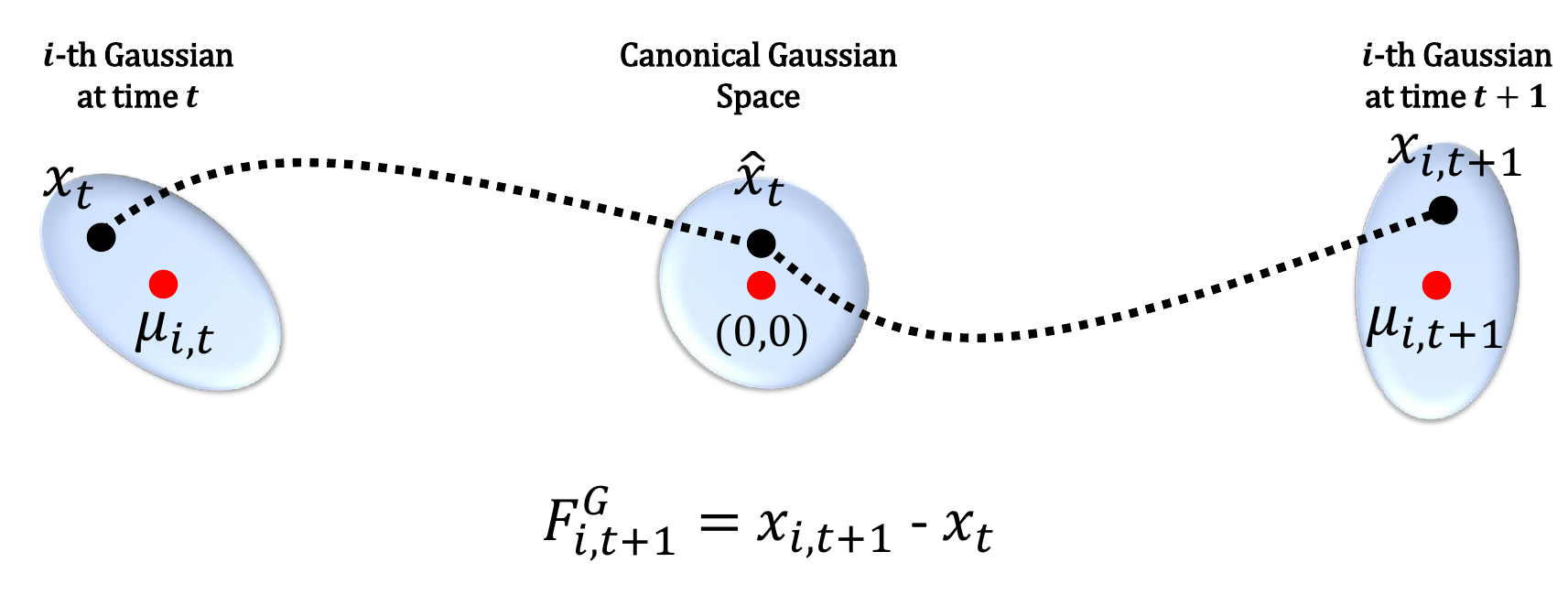}
   \caption{\textbf{The formulation of Gaussian flow.} We first project the point $x_t$ corresponding to the $i$-th Gaussian at time $t$ into the canonical Gaussian space, and then reproject this point from the canonical Gaussian space to the $i$-th Gaussian at time $t+1$.}
   \label{fig: gsflow}
\end{figure}

Motivated by~\cite{gao2024gaussianflow}, we formulate the Gaussian flow $F_{t \to t+1}^G$ to simulate the motion of dynamic object in the scene, as shown in~\Cref{fig: gsflow}
Specifically, Gaussian flow corresponds to the deformation of 3D Gaussians from time \( t \) to \( t+1 \) in the camera viewpoint \( C_{t+1} \). 
For point \( x_t \), we first transform it to the canonical space~\footnote{The canonical space mentioned here should not be confused with the canonical space in the Gaussian deformation field. It represents the standard Gaussian distribution space.} corresponding to the Gaussian at time \( t \):
\begin{equation}
    \hat{x}_t = \Sigma_{i,t}^{-1}(x_t-\mu_{i,t}),
\end{equation}
where $\mu_{i,t}$ and $\Sigma_{i,t}$ are the center position and covariance matrix of $i$-th Gaussian at the timestamp $t$.
Then we transform $\hat{x}_t$ back to the Gaussian at the next time step $t+1$:
\begin{equation}
    x_{i, t+1} = \Sigma_{i,t+1}\hat{x}_t+\mu_{i,t},
\end{equation}
where $\mu_{i,t+1}$ and $\Sigma_{i,t+1}$ are the center position and covariance matrix of $i$-th Gaussian at the timestamp $t+1$.
Therefore, the flow contribution from $i$-th Gaussian to this point can be defined as:
\begin{equation}
    F_{i, t \to t+1}^G = x_{i, t+1} - x_{t}.
\end{equation}
Finally, all Gaussian flow contributions to the point can be accumulated in a similar way to $\alpha$-blendering:
\begin{align}
    F_{t \to t+1}^G &= \sum_{i=1}^K w_i F_{i, t \to t+1}^G \\
                    &= \sum_{i=1}^K w_i (x_{i, t+1} - x_{t}) ,
\end{align}
where $w_i$ is the weight of $\alpha$-blendering.
Note that since the computation of forward optical flow is referenced to frame \( I_t \), the Gaussian flow should be consistently rendered under the camera viewpoint \( C_t \), corresponding to the decoupled motion flow. It represents the  2D splatting of the Gaussian deformation field from time \( t \) to \( t+1 \) when the camera viewpoint remains unchanged.

\subsection{More Implementation Details}
\label{appendix: details}

\paragraph{Datasets.}
The NeRF-DS dataset consists of eight stereo camera video sequences of daily scenes. These scenes contain high-speed moving high-gloss surface objects and changing camera poses, which pose challenges for dynamic scene modeling. 
The HyperNeRF dataset includes additional complications such as topological changes and inaccurate camera poses.
For the NeRF-DS dataset, we use the default resolution 480$\times$270 for all scenes for training and testing. 
We train the model using images from the left camera and test it on the right camera. 
For the HyperNeRF dataset, we select four sets of scenes in the vrig subset (3D Printer, Chicken, Broom, and Banana) for training and testing, with 2$\times$ downsampling resolution of $536\times960$.

\paragraph{Implementation details.}
To achieve the differentiable Gaussian flow and differentiable camera pose, we integrate the forward and backward processes into our rasterizer. 
To provide reliable optical flow, we choose GMFlow~\cite{xu2022gmflow} as the default optical flow network. 
In order to stabilize the initialization process of the scene, we introduce motion constraints only after the Gaussian starts to deform and move. 
In the NeRF-DS dataset, we set the weight of the flow loss $\lambda$ to 0.5; while in the HyperNeRF scene, it is set to 0.1. 
Camera pose optimization is also only activated during the Gaussian deformation stage. 

\paragraph{Data sampling mechanism.}
We adopt the same data sampling strategy as the baseline method, \ie reading image sequences in a randomly shuffled order. For an $N$-frame video, the frames are shuffled and then read sequentially. In each iteration, we read two frames and calculate the optical flow between them. To enhance efficiency, the second image from the last iteration is used as the first image in the current iteration. Thus, except for the first iteration, only one new image is read in each subsequent iteration. Consequently, there are $N-1$ iterations per epoch, with optical flow computed once in each iteration. This strategy balances the introduction of accurate motion priors with maintaining training efficiency.
During the first epoch of training, we calculate the optical flow for all adjacent frame pairs, resulting in a total of $N-1$ optical flow maps. In subsequent epochs, we do not reshuffle the image sequence, allowing us to reuse the optical flow maps calculated in the first epoch. This effectively eliminates the need to recompute optical flow maps in each epoch, significantly reducing computational overhead. 

\paragraph{Training time and GPU memory.} 
For model training, we list the training time per scene and peak memory usage on the NeRF-DS dataset as shown in~\Cref{tab: training_time,,tab: max_gpu_memory}, providing a comprehensive assessment of resource usage during training. Compared to our baseline, our approach incurs increased training time and peak memory usage. This is primarily due to the additional rendering of Gaussian flow and the refinement of camera poses, which are necessary for our method.

\begin{table}[ht]
\centering
\caption{\textbf{Training time comparison across different models.}}
\label{tab: training_time}
\resizebox{\textwidth}{!}{
\begin{tabular}{cccccccc}
    \toprule
    {Training Time}        & {As} & {Basin} & {Bell} & {Cup} & {Plate} & {Press} & {Sieve} \\ 
    \midrule
    Baseline             & 1h 1m       & 1h 11m         & 1h 42m        & 1h 3m        & 1h 0m          & 0h 51m         & 0h 57m         \\ 
    Ours (w/o pose refinement) & 1h 8m   & 1h 15m         & 1h 53m        & 1h 13m       & 1h 6m          & 0h 58m         & 1h 3m          \\ 
    Ours                 & 1h 33m      & 1h 46m         & 2h 4m         & 1h 34m       & 1h 30m         & 1h 17m         & 1h 25m         \\ 
    NeRF-DS              & 6h 43m      & 6h 48m         & 6h 49m        & 6h 50m       & 6h 53m         & 6h 48m         & 6h 47m         \\ 
    \bottomrule
\end{tabular}
}
\end{table}

\begin{table}[ht]
\centering
\caption{\textbf{Max GPU memory usage comparison across different models.}}
\label{tab: max_gpu_memory}
\begin{tabular}{cccccccc}
    \toprule
    {Max GPU Memory (GB)} & {As} & {Basin} & {Bell} & {Cup} & {Plate} & {Press} & {Sieve} \\ \midrule
    Baseline            & 15.67       & 13.61          & 15.97         & 15.29        & 9.66           & 10.65          & 12.17          \\ 
    Ours                & 16.61       & 14.52          & 17.73         & 15.70        & 10.62          & 11.62          & 12.97          \\ 
    \bottomrule
\end{tabular}
\end{table}

\paragraph{FPS, number of 3D Gaussians and storage.}  
We provide statistics of FPS and number of Gaussians on NeRF-DS dataset, as shown in~\Cref{tab: fps}. In most scenes of NeRF-DS, our method MotionGS achieves real-time rendering (FPS$>$30).

\begin{table}[ht]
\centering
\caption{\textbf{FPS, number of 3D Gaussians and storage on the NeRF-DS dataset per scene.}}
\label{tab: fps}
\setlength{\tabcolsep}{1.5em}{
\begin{tabular}{c|ccc}
\toprule
{Scene} & {FPS} & {Nums} & {Storage} \\
\midrule
AS         & 50   & 178K   & 49M \\
Basin      & 29   & 250K   & 70M \\
Bell       & 19   & 379K   & 97M \\
Cup        & 35   & 200K   & 54M \\
Plate      & 32   & 220K   & 58M \\
Press      & 45   & 197K   & 53M \\
Sieve      & 35   & 200K   & 54M \\
\bottomrule
\end{tabular}
}
\end{table}

\subsection{More Ablations}
\label{appendix: ablations}

We summarize the ablations on other choices of our proposed framework in~\Cref{tab:other-ablation}. For fair comparison, we do not activate the proposed camera pose refinement module during training, since it also influences the flow calculation. Our interpretation and analysis of the ablations are as follows.

\begin{table}[ht]
  \centering
  \caption{\textbf{Ablations on other choices of our proposed framework.}
  For fair comparison, we do not activate the proposed camera pose refinement module during training.
  }
  \resizebox{\linewidth}{!}{
    \begin{tabular}{c|l|ccc}
    \toprule
    Row & Setting & PSNR $\uparrow$ & SSIM $\uparrow$ & LPIPS $\downarrow$ \\
    \midrule
    1 & Baseline & 23.61  & 0.8394  & 0.1970  \\
    2 & w/o Motion mask & 23.13  & 0.8242  & 0.2249  \\
    3 & Different depth choice (Midas~\cite{Ranftl2022}) & 23.58  & 0.8384  & 0.1969  \\
    4 & Different optical flow network (FlowFormer~\cite{huang2022flowformer})& \cellcolor{yellow}23.97  & \cellcolor{yellow}0.8525  & 0.1893  \\
    5 & Different optical flow network (MDFlow~\cite{kong2022mdflow})& 23.25 & 0.8308 &	0.2137
  \\
    6 & Self-supervised flow supervision loss & 23.76  & 0.8474  & \cellcolor{yellow}0.1807 \\
    7 & Lower flow loss weight ($\lambda=0.2$) & 23.46  & 0.8343  & 0.2042  \\
    8 & Higher flow loss weight ($\lambda=0.8$) & 23.75  & 0.8470  & 0.1819  \\
    9 & Ours (w/o camera pose refinement)  & \cellcolor{pink}24.12 & \cellcolor{pink}0.8609 & \cellcolor{pink}0.1763 \\
    \bottomrule
    \end{tabular}%
    }
  \label{tab:other-ablation}%
\end{table}%

\paragraph{Effectiveness of motion mask (row 2).}
Introducing a motion mask allows the motion flow to focus on the motion of dynamic objects, thereby reducing interference from static areas. 
When the motion mask is removed, the performance declines. We attribute this degradation to inaccurate optical flow in the background areas, which introduces errors in the motion guidance and subsequently leads to incorrect Gaussian deformations.

\begin{figure}[t]
   \centering
   \includegraphics[width=0.8\linewidth]{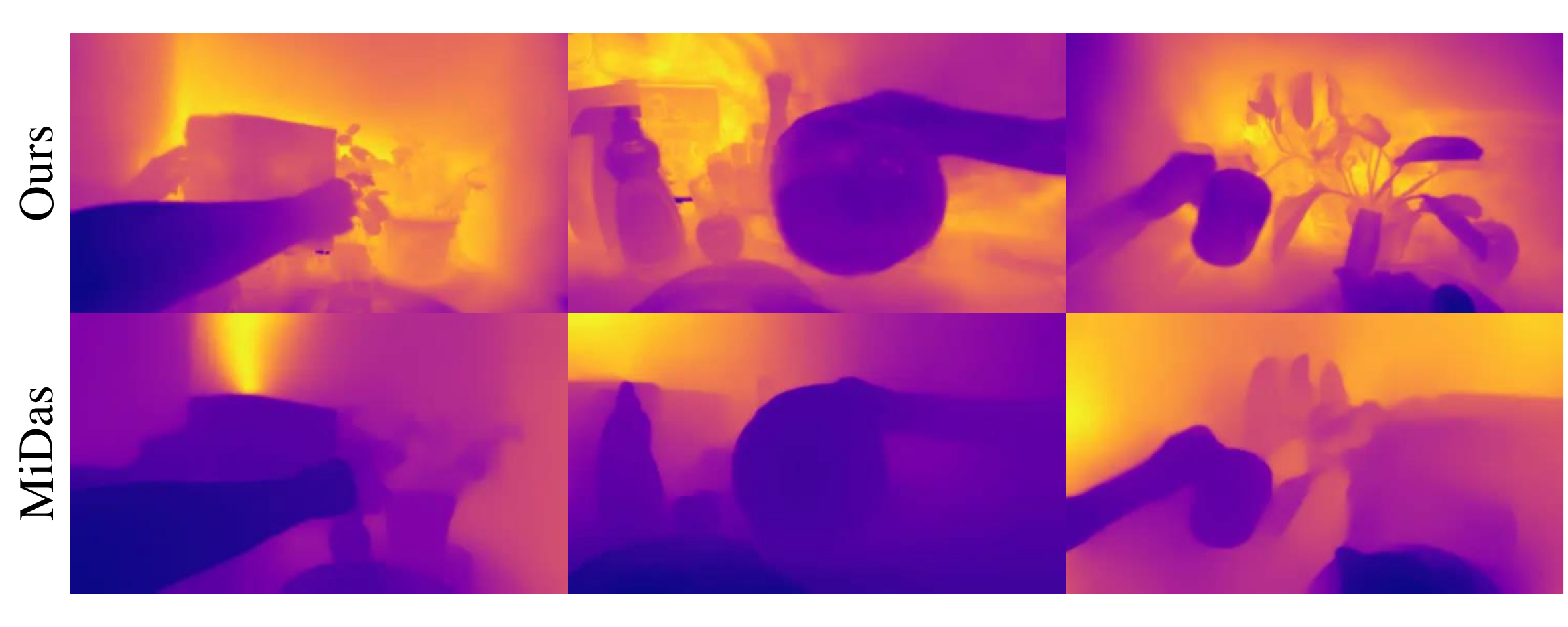}
   \caption{\textbf{Rendered depth from 3D Gaussian splatting (ours) and off-the-shelf monocular depth estimator (MiDas).} Our rendered depth has richer details and is scale-aligned with the scene. MiDas rendered depth is usually more smooth and suffers from scale ambiguity.
   }
   \label{fig: depth}
\end{figure}

\paragraph{Different depth choice (row 3).}
Estimating the accurate depth maps for depth warping is a critical issue when calculating camera flow. We find that using depth prediction from offline estimator Midas~\cite{Ranftl2022} yields suboptimal results. This approach degrades the quality of subsequent motion flow, reducing the accuracy of motion constraints and ultimately impacting the reconstruction quality. 
We attribute this degradation to the inherent scale ambiguity~\cite{zhu2024scaledepth} in the depth estimator, as shown in~\Cref{fig: depth}. 
In contrast, using rendered depth by 3DGS ensures scale and geometric consistency and provides superior detail.

\paragraph{Different optical flow network (row 4-5).}
Our method relies on existing 2D optical flow estimators to provide motion guidance for the 3D Gaussian fields. The choice of optical flow prior can lead to performance differences. When we replace GMFlow~\cite{xu2022gmflow} with another supervised method FlowFormer~\cite{huang2022flowformer}, the performance deteriorates. This is mainly due to the fact that FlowFormer performs inadequately in the "plate" scene, resulting in an overall performance decrease.
Additionally, when we replace GMFlow~\cite{xu2022gmflow} with a self-supervised method MDFlow~\cite{kong2022mdflow}, the performance is even worse.
This phenomenon may also illustrates the importance of accurate motion priors, while erroneous or noisy motion constraints may even have a negative effect on the optimization.

\paragraph{Self-supervised flow supervision loss (row 6).}
Inspired by self-supervised optical flow estimation methods, we attempt to provide motion priors in a self-supervised manner. Specifically, we estimate the Gaussian flow corresponding to the optical flow and use it to warp the $I_t$ frame. We then compute the photometric loss with the $I_{t+1}$ frame. As shown in the table, this method outperforms our baseline but is less effective compared to our proposed method.
We hypothesize that the discrepancy arises because the self-supervised loss may not provide accurate supervision in areas with similar colors. Nevertheless, it is evident that employing self-supervised optical flow loss can reduce dependence on off-the-shelf optical flow estimation. When an optical flow estimation network is either unavailable or inaccurate, this approach can serve as a valuable alternative to improve rendering quality.

\paragraph{Different flow loss weights (row 7-8).}
We compare the rendering performance under different flow loss weights. The results indicate that the selected weight ($\lambda=0.5$) achieves the best rendering quality. 
We speculate that excessively large loss weights may disrupt the original optimization process based on rendering losses, while too small weights may result in insufficient motion guidance.

\subsection{More Visualizations}
\label{appendix: visualizations}

\begin{figure}[htbp]
   \centering
   \includegraphics[width=\linewidth]{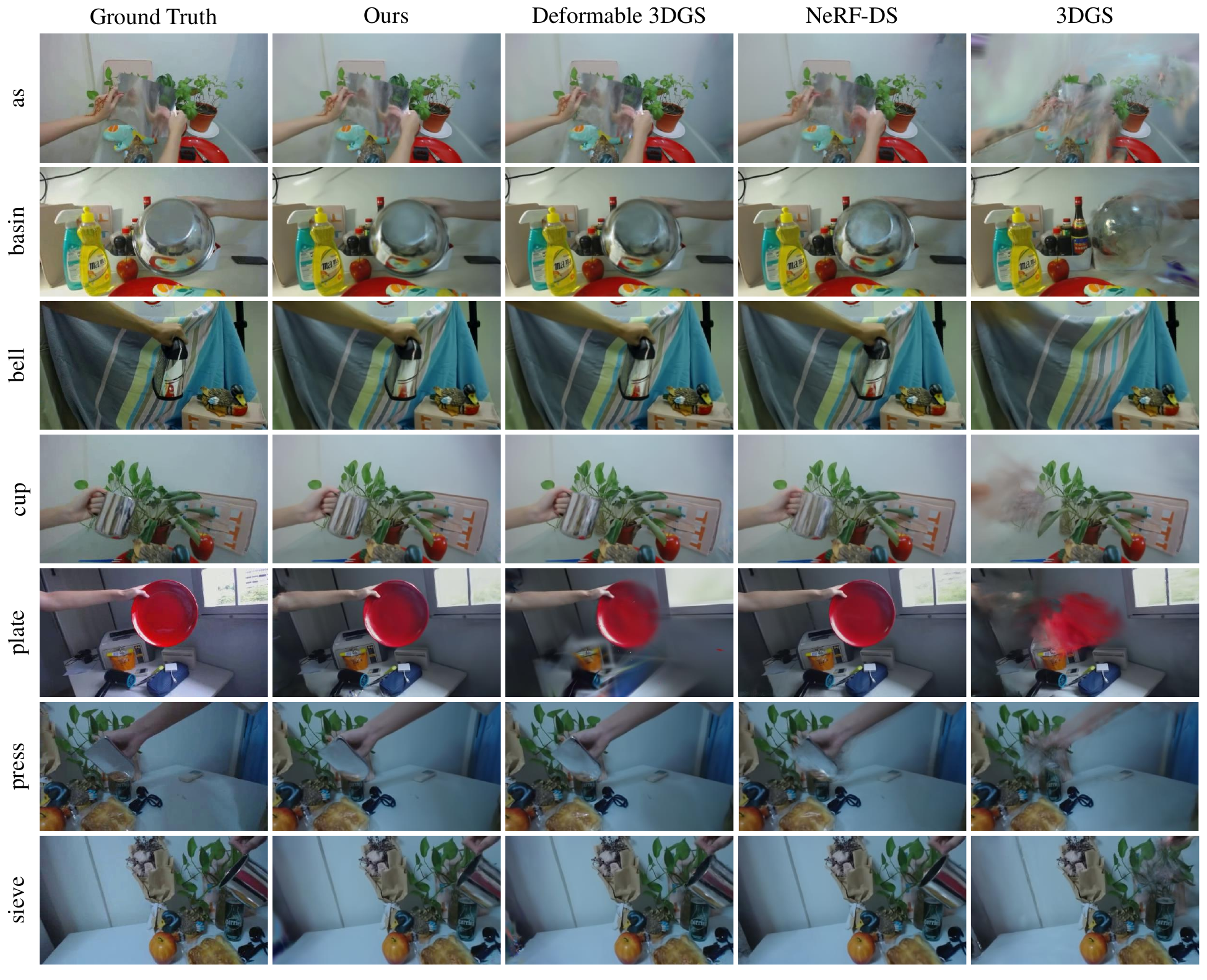}
   \vspace{-1em}
   \caption{\textbf{Qualitative comparison on NeRF-DS dataset per-scene.} Compared with the state-of-the-art methods, our method can render more reasonable details, especially on dynamic objects.}
   \label{fig: nerfds_full}
\end{figure}

\begin{figure}[htbp]
   \centering
   \includegraphics[width=\linewidth]{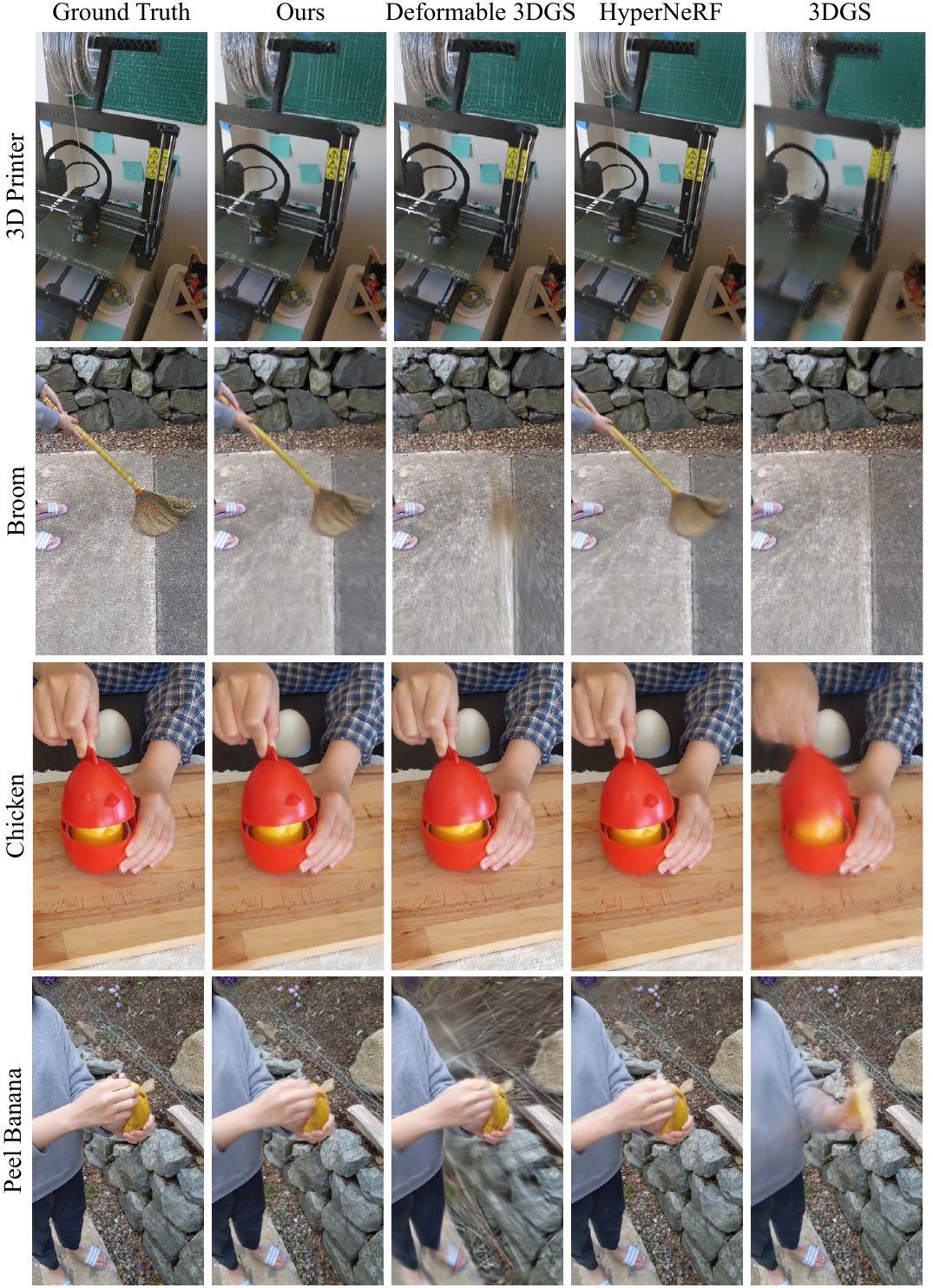}
   \vspace{-1em}
   \caption{\textbf{Qualitative comparison on HyperNeRF dataset per-scene.} Compared with the state-of-the-art methods, our method is more robust in reconstructing dynamic scenes. Even if the input camera pose is not accurate on HyperNeRF dataset, our method can adaptively optimize the camera poses and produce reasonable rendering results.}
   \label{fig: hyper_full}
\end{figure}

\begin{figure}[htbp]
   \centering
   \includegraphics[width=0.9\linewidth]{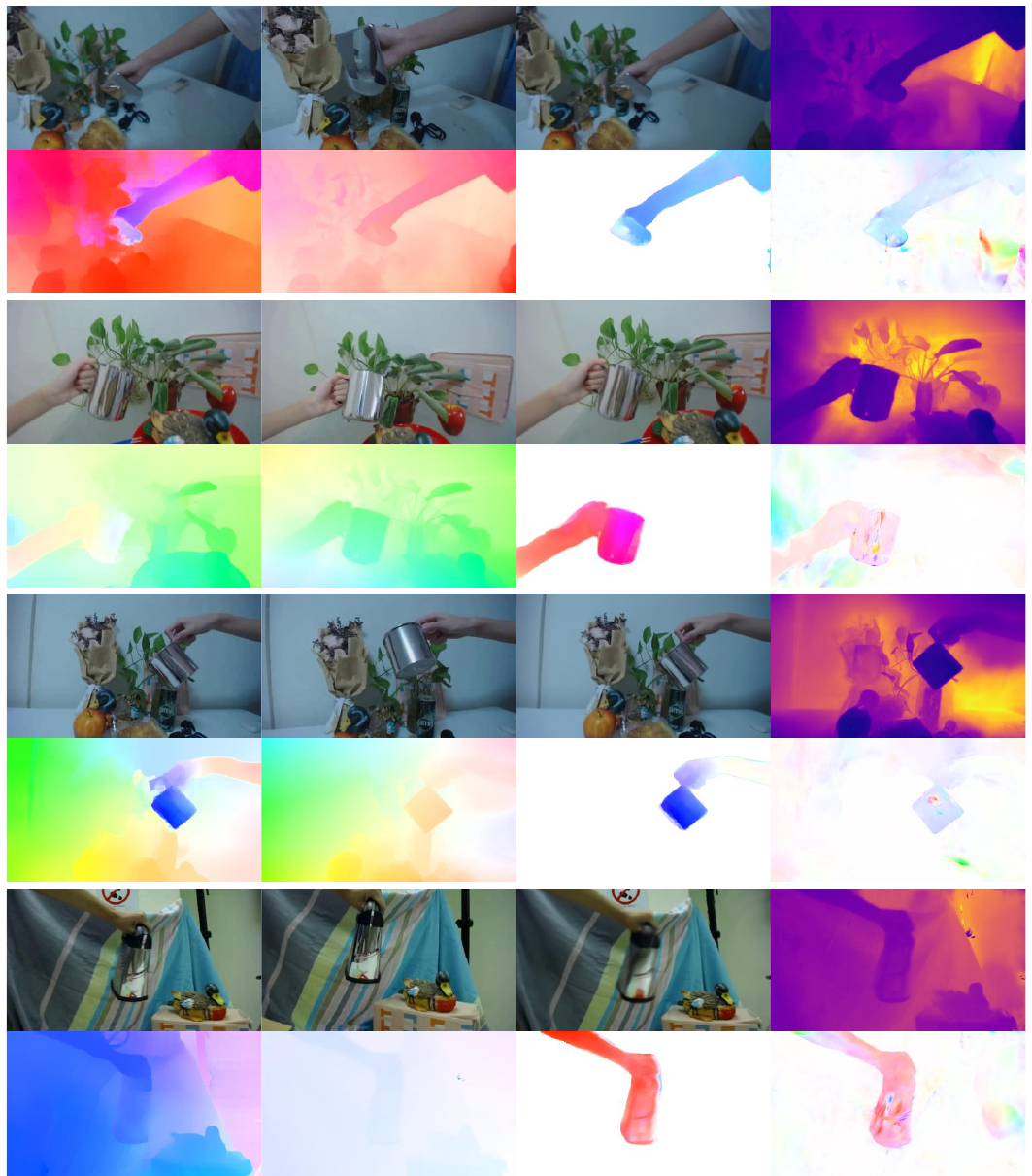}
   \caption{\textbf{Visualization of all data flows.} In order:  ground truth of $I_t$, ground truth of $I_{t+1}$, rendered image of $I_t$, rendered depth of frame $I_t$, optical flow, camera flow, motion flow, Gaussian flow.}
   \label{fig: example_full}
\end{figure}

Please refer to~\Cref{fig: nerfds_full,,fig: hyper_full,,fig: example_full}.

\subsection{Limitation}
\label{appendix: failure cases}

\begin{figure}[htbp]
   \centering
   \includegraphics[width=\linewidth]{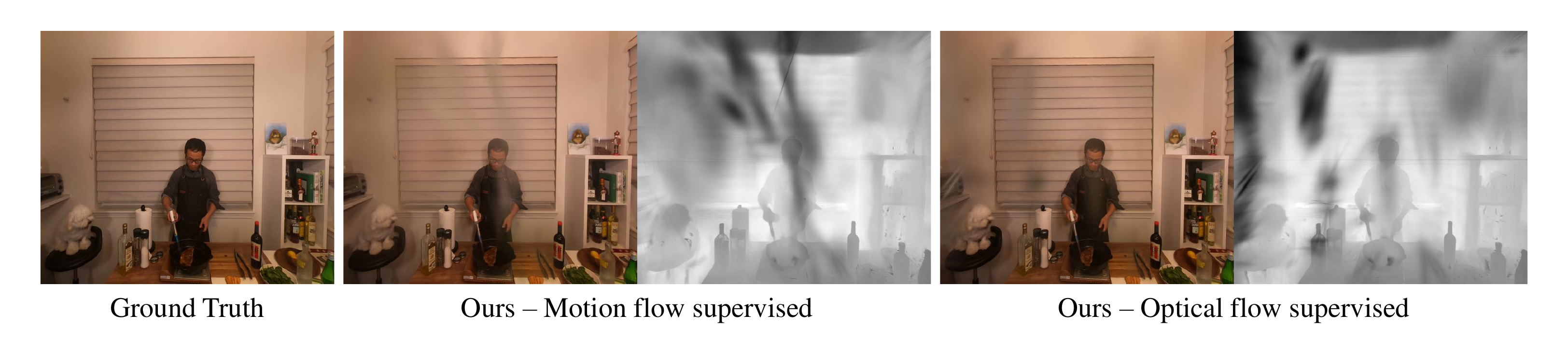}
   \caption{\textbf{Failure case in DyNeRF dataset.} 
    Since the viewpoints are fixed and sparse, neither motion flow nor optical flow can help our method avoid floating artifacts.
   }
   \label{fig: fail}
\end{figure}

During our experiments, we identify several unresolved issues. Specifically, when applying our method to the DyNeRF dataset~\cite{li2022neural}, we encounter significant challenges as illustrated in~\Cref{fig: fail}. 
Upon further analysis, we find that the fixed and sparse camera viewpoints in the DyNeRF dataset hinder accurate depth rendering, affecting subsequent camera flow calculations and leading to artifacts. The inaccuracies in motion flow primarily comes from the inaccuracy of the camera flow, rather than a failure of the optical flow estimation itself. It is also important to clarify that the DyNeRF dataset is not continuous monocular video but rather dynamic scenes with sparse viewpoints, which posed challenges to the canonical 3D Gaussian initialization.
Moving forward, our focus will be on addressing these issues to further improve the robustness of our model in dynamic scene reconstruction. We aim to develop more stable and reliable motion priors and adapt our approach to handle scenarios with minimal object movement more effectively. By doing so, we hope to extend the applicability and reliability of our method across a wider range of dynamic scenes.

\subsection{Broader Impacts}
\label{appendix: social impacts}

To the best of our knowledge, the proposed method will not have significant negative social impact. The proposed dynamic reconstruction method can be used to reconstruct and render some daily dynamic scenes. Users can use the video shot by their mobile phones as input to obtain an explicit 3D asset represented by a 3D Gaussian and a deformation field. This 3D asset can be used for subsequent editing, development, secondary creation for entertainment.

\subsection{Data Availability}
\label{appendix: data}

The datasets that support the findings of this study are available in the following repositories: 
NeRF-DS~\cite{yan2023nerf} at \url{https://github.com/JokerYan/NeRF-DS/releases/tag/v0.1-pre-release} under Apache-2.0 license,
HyperNeRF~\cite{park2021hypernerf} at \url{https://github.com/google/hypernerf/releases/tag/v0.1} under Apache-2.0 license.
The code of our baseline~\cite{yang2023deformable} is available at \url{https://github.com/ingra14m/Deformable-3D-Gaussians} under MIT license.

\vspace{5em}

\clearpage

\end{document}